\documentclass{article}

\PassOptionsToPackage{numbers, compress}{natbib}
\usepackage[preprint]{neurips_2025}

\usepackage[utf8]{inputenc} % allow utf-8 input
\usepackage[T1]{fontenc}    % use 8-bit T1 fonts
\usepackage{hyperref}       % hyperlinks
\usepackage{url}            % simple URL typesetting
\usepackage{booktabs}       % professional-quality tables
\usepackage{amsfonts}       % blackboard math symbols
\usepackage{nicefrac}       % compact symbols for 1/2, etc.
\usepackage{microtype}      % microtypography
\usepackage{xcolor}         % colors
\usepackage{xspace, overpic, mdframed, setspace, enumitem, wrapfig, multirow, amsmath, caption}
\usepackage{algorithm, algorithmic}
\usepackage[capitalize,noabbrev]{cleveref}

\title{Addressing the ID-Matching Challenge in Long Video Captioning}

\author{Zhantao Yang$^{1,2}$$^\star$, Huangji Wang$^{1}$$^{\star}$, Ruili Feng$^{2}$, Han Zhang$^{1,2}$, \\ Yuting Hu$^{1}$, 
    Shangwen Zhu$^{1}$, Junyan Li$^{1}$, Yu Liu$^{2}$, Fan Cheng$^{1}$$^\dagger$ \\ $^1$Shanghai Jiao Tong University, $^2$Alibaba group \\
    \footnotesize\texttt{\{ztyang196, chengfan85\}@gmail.com}}
\newcommand{\methodabbr}{\textsc{RICE}\xspace}

\newcommand{\Method}{\textbf{R}ecognizing \textbf{I}dentities for \textbf{C}aptioning \textbf{E}ffectively\xspace}
\newcommand{\task}{ID-Matching\xspace}
\newcommand{\benchmark}{\textsc{RICE}-benchmark\xspace}

\newcommand{\vlm}{\textsc{RICE}-8B\xspace}
\newcommand{\tocite}[1]{\textcolor{red}{[TO CITE]}}

\newmdenv[%
    linewidth=1pt,       % Line thickness
    topline=false,       % Optional: Disable top line
    bottomline=false,    % Optional: Disable bottom line
    skipabove=\baselineskip,
    skipbelow=\baselineskip,
    innertopmargin=0.5\baselineskip,
    innerbottommargin=0.5\baselineskip,
    innerleftmargin=0.5cm,
    innerrightmargin=0.5cm,
]{examplebox}

\begin{document}

\maketitle

\begin{abstract}

Generating captions for long and complex videos is both critical and challenging, with significant implications for the growing fields of text-to-video generation and multi-modal understanding. One key challenge in long video captioning is accurately recognizing the same individuals who appear in different frames, which we refer to as the \textbf{\task} problem. Few prior works have focused on this important issue. Those that have, usually suffer from limited generalization and depend on point-wise matching, which limits their overall effectiveness. 
In this paper, unlike previous approaches, we build upon LVLMs to leverage their powerful priors. We aim to unlock the inherent \task capabilities within LVLMs themselves to enhance the \task performance of captions.
Specifically, we first introduce a new benchmark for assessing the \task capabilities of video captions. Using this benchmark, we investigate LVLMs containing GPT-4o, revealing key insights that the performance of \task can be improved through two methods: 1) enhancing the usage of image information and 2) increasing the quantity of information of individual descriptions.
Based on these insights, we propose a novel video captioning method called \Method (\methodabbr).
Extensive experiments including assessments of caption quality and \task performance, demonstrate the superiority of our approach.
Notably, when implemented on GPT-4o, our \methodabbr improves the precision of \task from 50\% to 90\% and improves the recall of \task from 15\% to 80\% compared to baseline. \methodabbr makes it possible to continuously track different individuals in the captions of long videos.

\end{abstract} 
\section{Introduction}\label{sec:introduction}

Generating captions for long and complex videos has significant implications for the rapidly evolving fields of text-to-video generation~\citep{singer2022make,hong2022cogvideo,wu2023tune,zhang2023controlvideo,li2018video,du2024learning,menapace2024snap,wang2024recipe} and multi-modal understanding~\citep{videollava-7b,sun2024generative,zhou2024streaming,islam2024video,chen2024panda,ventura2024covr,huang2024vtimellm,wang2024mitigating,nguyen2024improving,mei2024wavcaps}. However, long video captioning poses significant challenges, one of the primary ones being the accurate identification of the same individuals who appear repeatedly across multiple frames, referred to as the \task problem. 
If IDs are misrecognized, actions of the same person may be mistaken for different individuals, or actions of different people attributed to one, causing significant video understanding errors.
Despite its importance, this issue has received relatively little attention in previous methodologies.

\begin{figure*}[t]
    \centering
    \includegraphics[width=\textwidth]{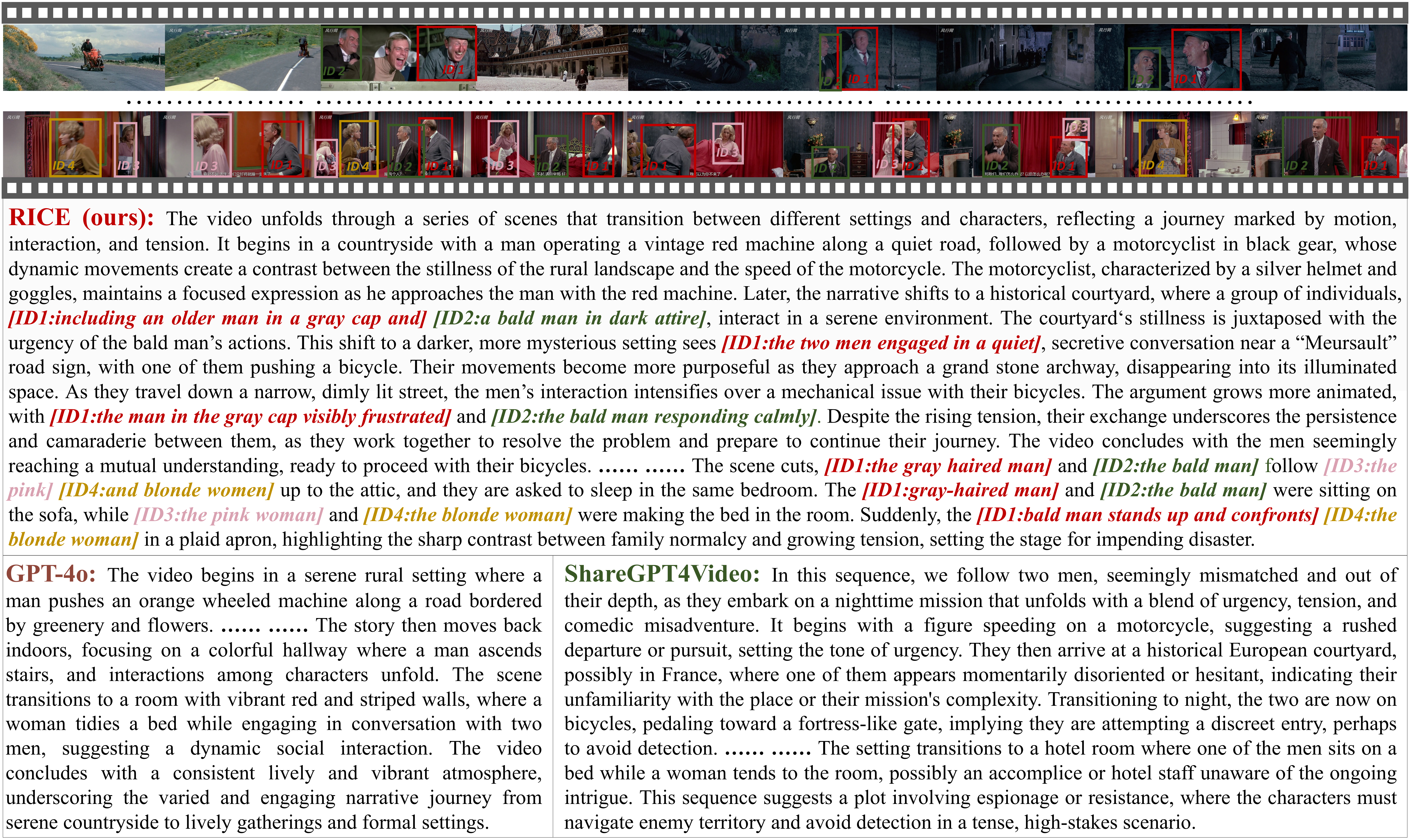}
    \vspace{-15pt}
    \caption{
    An example of video captions for a video \textbf{lasting around 480 seconds (around 200 key frames)} obtained by: GPT-4o, ShareGPT4Video, and \methodabbr. For simplicity, we have left out some intermediate key frames and their captions. The results show that RICE generates more detailed and accurate captions while \textbf{keeping tracking the main individuals throughout the long video} (The repeatedly appearing individuals are highlighted in different colors.) In contrast, GPT-4o and ShareGPT4Video lose track of key individuals over time, misinterpreting content about one person as involving multiple people.
    }
    \vspace{-10pt}
    \label{fig:caption_example}
\end{figure*}

Previous match-based methods, including face recognition~\citep{kotwal2025review,dominguez2024metrics, deandres2024frcsyn, wen2023divide} and person re-identification~\citep{ye2024transformer, jiang2025attributes, wang2024large, joseph2025clothes, jiang2025laboratory}, suffer from many problems when used for \task in long video captioning: 1) They are usually trained on specific datasets and fail in diverse scenarios—for example, person re-ID relies on surveillance viewpoints, while face recognition needs frontal faces within a certain range of distances to the camera. 2) They focus on different body regions (head and shoulders vs. full body), making integration difficult when matching frames containing people with different body regions. 3) They also require extra preprocessing and additional complex pipelines to generate captions. Other methods~\citep{park2020identity, han2023autoad, han2023autoad, ji2024ida} train on limited data with simplified settings, matching only a few frames and choosing ID from limited options. Finally, all mentioned previous methods rely on point-wise matching rather than predicting a consistent sequence required in long video captioning. Any hallucination may lead to inevitable conflicts—for instance, if frame A matches B, and A matches C, but B does not match C, it becomes impossible to form a coherent sequence.

We argue that current large vision language models (LVLMs) already exhibit strong image understanding and generalization, effectively handling diverse scenes and multiple people in video captioning. This paper aims to tap into the inherent \task capabilities within LVLMs, and leverage their rich knowledge to generate captions that are both high-quality and accurate in \task.
However, two main challenges remain: \textbf{1)} First, there is a lack of benchmarks to study this problem. Most videos in existing datasets either do not contain scenes with multiple people appearing repeatedly~\citep{Youcook2} or lack annotations indicating which frames each person appears in~\citep{caba2015activitynet,song2024moviechat}. Additionally, there is a lack of methods to extract predicted ID sequences from the captions, and metrics to evaluate the \task performance. 
\textbf{2)} Second, processing long videos leads to attention dilution and increased noise~\citep{liu2024exposing}, making it difficult to associate the same IDs across distant frames through attention mechanisms. Previous methods often split long videos into segments, caption each segment, and then summarize them~\citep{chen2024sharegpt4video}. However, their summarization usually relies on plain text, which easily loses ID information.
Previous methods often cut long videos into segments to address this issue~\citep{chen2024sharegpt4video}. They caption each segment and then summarize them for a final caption. However, their summarization stage typically solely uses text, which can easily lose ID information.

To address this problem, we first develop a set of tools to explore the \task problem, resulting in \textbf{\benchmark}. This includes: \textbf{1)} the first dataset specifically collected for studying the \task problem in long video captioning; \textbf{2)} a method for automatically extracting predicted ID sequences from the captions; and \textbf{3)} several metrics to evaluate \task performance. 
Next, we use \benchmark to analyze GPT-4o~\citep{gpt4o} and find that: \textbf{1)} the captioning stage can be improved by enhancing the usage of image information, achieved by captioning multiple frames within a single dialogue. \textbf{2)} increasing the quantity of information describing each individual—by incorporating more features in descriptions—significantly improves textual similarity among matched individuals during summarization, thus reducing ID information loss and improving \task performance of the final captions. Based on these insights, we propose \Method (\methodabbr) to enhance \task performance while preserving caption quality.

As shown in \cref{fig:caption_example}, \methodabbr can correctly recognize individuals who appear repeatedly in long videos (around 480 seconds and over 100 key frames) while maintaining high-quality captions. We conduct extensive experiments for evaluation, using multiple metrics to assess caption quality and \task performance. The experimental results demonstrate that \methodabbr outperforms the baseline and the previous state-of-the-art, ShareGPT4Video~\citep{chen2024sharegpt4video}, across all evaluated benchmarks. Remarkably, when implemented on GPT-4o, our \methodabbr significantly improves the precision of \task \textbf{from 50\% to 90\%} and improves the recall of \task \textbf{from 15\% to 80\%} compared to the baseline. 
% Furthermore, we leverage the \task capabilities of GPT-4o to track long videos with shot changes, using the bounding boxes from the results as a supplementary context for the captions. 
We hope our work offers valuable insights and tools to the field of video captioning.
\section{Related works}\label{sec:relatedworks}

% \begin{figure}[t]
% \centering
% \begin{overpic}[width=\linewidth]{images/datasets.pdf}
% \end{overpic}
% \vspace{-16pt}
% \caption{An example from the dataset in \benchmark.}
% \vspace{-8pt}
% \label{fig:id-matching-dataset}
% \end{figure}

Video captioning is a pivotal research topic in multi-modal understanding. Early methods focus on identifying objects within the video and assigning terminology to them, ultimately merging these elements into a caption~\citep{6751448, Kojima2002}. They typically evaluate captions using similarity scores, including BLEU~\citep{papineni2002bleu}, METEOR~\citep{denkowski2014meteor}, ROUGE~\citep{lin2004rouge}, CIDEr~\citep{vedantam2015cider}, and SPICE~\citep{anderson2016spice}. However, the resulting captions are often overly concise, failing to meet the demands of complex multi-modal tasks.

\begin{figure*}[t]
    \centering
    \includegraphics[width=\textwidth]{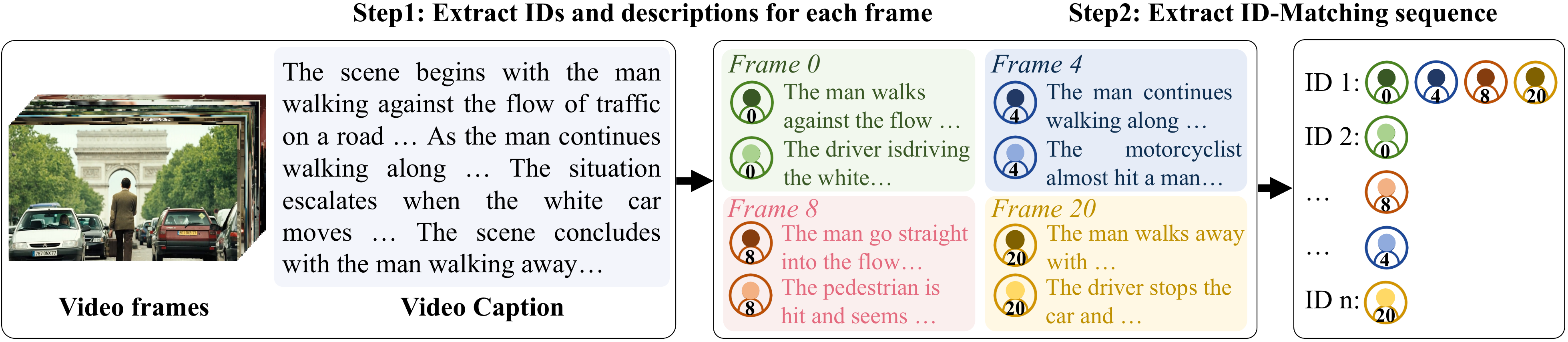}
    \vspace{-15pt}
    \caption{The method for automatically extracting the predicted \task sequence from the caption involves two steps: 1) Given the video frames and the caption, extract the IDs and their corresponding descriptions for each frame; 2) Extract the \task sequence by sequentially comparing the descriptions of the IDs in the current frame with those in the previous frame to determine whether they are the same ID. Details about the method can be found in \cref{subsec:app_extract_method}.}
    \vspace{-5pt}
    \label{fig:evaluation_pipeline}
\end{figure*}

To address this issue and generate more detailed descriptions, dense captioning methods have emerged~\citep{krishna2017dense, yang2023vid2seq, zhou2024streaming, qin2024question}, which usually rely on manually annotated datasets for training, such as ActivityNet~\citep{caba2015activitynet}, Didemo~\citep{anne2017localizing}, MovieChat~\citep{song2024moviechat}, YouCook2~\citep{Youcook2}, Youku-mPLUG~\citep{xu2023youku}, ViTT~\citep{vitthuang2020multimodal}, MSRVTT~\citep{xu2016msr}, MSVD~\citep{vaish2017color}, SoccerNet~\citep{giancola2018soccernet}, and SPORTS-MOT~\citep{cui2023sportsmot}. Some datasets directly label complex text descriptions, while others segment videos into key clips, with each segment accompanied by a brief description. However, manual annotation limits the coverage of video content. Recently, LVLMs have helped overcome this limitation. For example, ShareGPT4Video~\citep{chen2024sharegpt4video} uses GPT-4V to extract key frames and generate detailed, accurate descriptions, which are then combined into a comprehensive caption for the entire video. Despite their strengths, these methods still face the challenge of the \task problem. 

Matching methods such as face recognition~\citep{kotwal2025review,dominguez2024metrics, deandres2024frcsyn, wen2023divide} and person re-identification~\citep{ye2024transformer, jiang2025attributes, wang2024large, joseph2025clothes, jiang2025laboratory} struggle with the \task problem in long video captioning due to limited generalization and reliance on point-wise matching. Other approaches~\citep{park2020identity, han2023autoad, han2023autoad, ji2024ida} train on limited data with simplified settings, matching only a few frames and selecting IDs from a small pool, also facing point-wise matching challenges. In contrast, this paper explores the capabilities of \task in LVLMs to enhance long video captioning performance.
\section{Tools for evaluating \task performance of video captions}\label{sec:benchmark}

The lack of specialized tools significantly hinders the exploration of the \task problem in video captioning. To fill this gap, we create a set of tools called \benchmark, which includes: \textbf{1) a dataset} that annotates the appearing individuals along with the frame indices in which they appear; \textbf{2) an extraction method} to automatically extract the predicted sequence of appearing frame indices for all individuals from the captions; and \textbf{3) evaluation metrics} to assess \task performance based on the predicted sequences and ground truth. To the best of our knowledge, this is the first benchmark of its kind.

\subsection{Dataset}\label{subsubsec:dataset}

The dataset comprises 374 videos randomly selected from multiple sources, including MovieChat~\citep{song2023moviechat}, Miradata~\citep{ju2024miradata} and Panda-70M~\citep{chen2024panda}, to ensure data diversity. These videos contain multiple roles that appear repeatedly across different frames. For each video, we first extract key frames using the method described in~\citep{chen2024sharegpt4video} and then select clips, removing those that do not contain multiple characters as well as instances where a single character appears continuously, as such cases are overly simple. Annotators manually identify all roles, record the frame indices in which each role appears, and perform a double-check to ensure accuracy. For detailed statistical information and construction details of the dataset, please refer to \cref{subsec:app_collecting_data} and the supplementary materials.
% The resulting dataset includes 653 unique individuals, with each person appearing an average of 4 times in the clips, which have an average length of 25 frames. 
% The distribution of the length of the intervals between consecutive appearances of the same person is shown in \td{figure}(b).

\subsection{Method to extract predctions from captions}\label{subsubsec:extract_method}

Assessing the \task performance of a caption requires extracting the predicted \task sequences from the caption, represented as $\{I_p^1, I_p^2, \dots, I_p^n\}$, containing $n$ distinct IDs. Each ID corresponds to a sequence indicating the indices of the frames in which it appears, denoted as $I_p^i = \{f_{i1}, f_{i2}, \dots, f_{ik}\}$, with $f_{ij}$ representing the index of the $j$-th frame where the $i$-th ID appears. However, manually extracting these frame index sequences for all individuals is time-consuming. To address this, we develop an automated pipeline leveraging GPT-4o, which achieves over 90\% accuracy compared to manual extraction.
As illustrated in \cref{fig:evaluation_pipeline}, the method comprises two steps: 1) extracting all IDs and their corresponding descriptions for each frame based on the frames and the caption; and 2) extracting the ID-matching sequence by sequentially comparing the descriptions of IDs in the current frame with those in the previous frame to determine whether they are the same ID.
Further details of the method, including all prompts used, are provided in \cref{subsec:app_extract_method}.

% \begin{table}[t]
% \centering
% \vspace{-4pt}
% \caption{
% List of abbreviations used in this paper.
% }
% % \vspace{2pt}
% \label{tab:abbreviation}
% \renewcommand{\arraystretch}{0.7}
% \resizebox{\linewidth}{!}{
%     \input{tables/abbreviations}
% }
% \vspace{-16pt}
% \end{table}

\begin{figure*}[t]
\centering
\includegraphics[width=\textwidth]{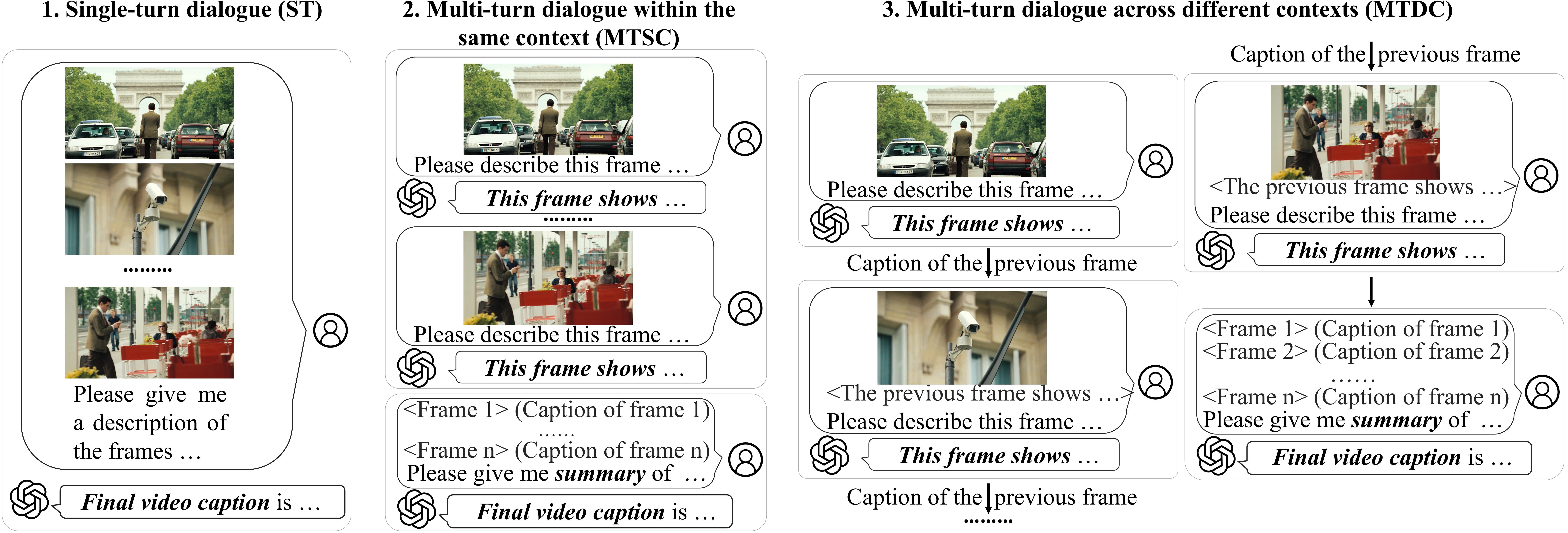}
\vspace{-15pt}
\caption{The three fundamental methods of LVLMs for annotating multiple frames, including ST, MTSC, and MTDC (see detailed definition in \cref{subsec:fundamental_methods}).}
\vspace{-10pt}
\label{fig:basic-model}
\end{figure*}

\subsection{Evaluation metrics for \task}\label{subsubsec:eval_metrics}

While an incorrect \task operation can lead to significant semantic errors, it often manifests as a discrepancy of only a single word—typically the subject of a specific action. Such subtle discrepancies render traditional similarity-based metrics ineffective.
To better evaluate \task performance, we employ three metrics: \textbf{1) Precision and 2) Recall} measure pairwise recognition accuracy. To compute these, we first decompose each predicted sequence $I_p^i$ and each ground truth sequence $I_g^i$ into multiple pairwise tuples, aggregating them into the set $\{(f_{ij}, f_{il})|1 \leq i \leq n, 1\le j < l\le k\}$, where $n$ is the number of IDs and $k$ is the number of frames containing the $i$-th ID. Precision and recall are then calculated as the ratio of correctly predicted tuples to the total predicted tuples and total ground truth tuples, respectively. \textbf{3)
Sequence similarity} evaluates the overall similarity between the predicted and ground truth sequences. We model this as a bipartite graph $G=(U,V)$, where $U$ represents the set of predicted sequences and $V$ the ground truth sequences. Each node 
$U_i$ corresponds to $I_p^i$. The edge weight between nodes $U_i$ and $V_j$ is defined as the length of the intersection between $I_p^i$ and $I_g^j$. Using the Hungarian algorithm~\citep{kuhn1955hungarian}, we find the optimal matching between predicted and ground truth sequences. The sequence similarity is then computed as the relative sum of matched edge weights normalized by the total weight.
\section{Enhancing \task in video captioning}\label{sec:analysis}

In this section, we introduce our method to enhance \task in video captioning. Specifically, we decompose the captioning process of LVLMs for long videos into three fundamental methods (detailed in \cref{subsec:fundamental_methods}). We then design comprehensive experiments to investigate these fundamental methods, from which we derive two key insights: enhancing image information usage (detailed in \cref{subsec:distance}) and increasing the amount of information in individual descriptions (detailed in \cref{subsec:similar_attributes}). Finally, in \cref{subsec:our_method}, based on these observations, we propose \Method(\methodabbr), a method implemented on LVLMs to enhance \task performance in long video captioning.

% In this section, we first introduce the three fundamental methods in video captioning in \cref{subsec:fundamental_methods} as the preliminary of analysis. Then, we separately introduce the two key insights we observe on GPT-4o using \benchmark, including enhancing image information usage (\cref{subsec:distance}) and improving the quantity of information of individual descriptions (\cref{subsec:similar_attributes}).

\subsection{Three fundamental methods of using LVLMs for video captioning}\label{subsec:fundamental_methods}
For LVLMs like GPT-4o, there are three fundamental methods to generate captions from multiple frames, distinguished by whether images are provided within the same dialogue or context. As illustrated in \cref{fig:basic-model}, these methods are:
\textbf{1) Single-Turn dialogue (ST):} All frames are provided simultaneously in a single dialogue. ST involves only the captioning stage.
\textbf{2) Multi-Turn dialogue within the Same Context (MTSC):} Captions are generated through multiple dialogues within the same context. Each dialogue contains a single image, and images from previous dialogues is accessible via history. MTSC has two sub-modes: with text (where captions from prior frames are included in instructions) and without text.
\textbf{3) Multi-Turn dialogue across Different Contexts (MTDC):} Captions are generated through multiple dialogues across different contexts. In this mode, image information from previous frames is not accessible, while textual information can still be provided via prompts.
Unlike ST, both MTSC and MTDC require an additional summarization stage to produce the final caption based on individual frame captions. Most existing methods can be seen as combinations or variations of these three fundamental methods.

\subsection{Enhancing the usage of visual information}\label{subsec:distance}

Images provide abundant and valuable visual information that can significantly aid in \task. Among the three fundamental methods defined in \cref{subsec:fundamental_methods}, both ST and MTSC leverage image information. In ST, multiple images are input simultaneously within a single dialogue, enabling direct attention operations across these images and thus fully utilizing the visual information. In contrast, MTSC can only access prior image information indirectly through the history when annotating the current frame. Additionally, the presence of numerous intervening text tokens—including prompt and response tokens—between the current image and previous image in history substantially increases the token distance between earlier and current images, which impedes effective attention computation. Therefore, ST is able to utilize image information more effectively than MTSC.
As for MTDC, it performs \task solely based on textual information, without incorporating visual information.

\begin{figure*}[t]
% \hspace{2pt}
\begin{minipage}[c]{0.75\textwidth}
\centering
\begin{overpic}[width=0.99\linewidth]{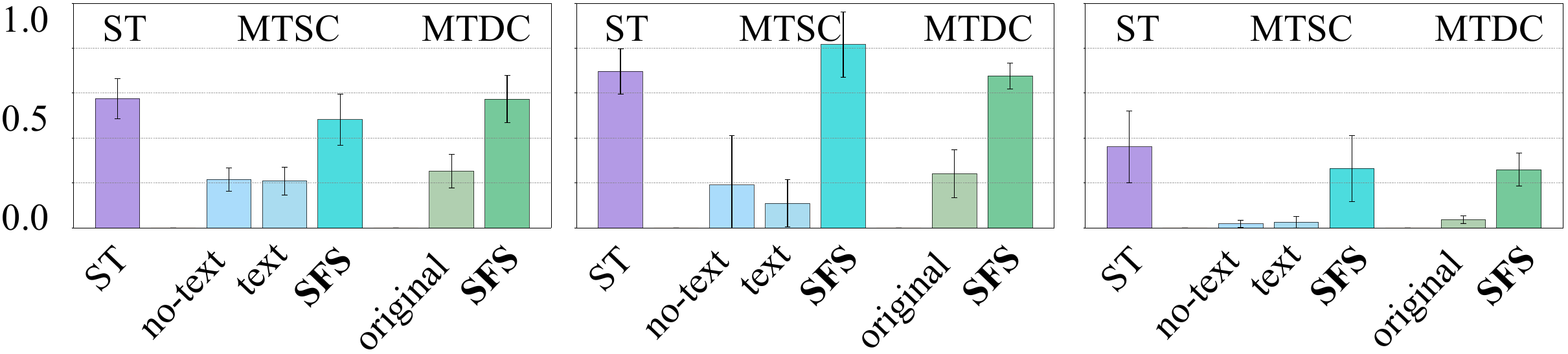}
\put(4,-3){\small (a) Sequence similarity}
\put(44,-3){\small (b) Precision}
\put(78,-3){\small (c) Recall}
\end{overpic}
\vspace{4pt}
\caption{Comparison of ST, MTSC, and MTDC (introduced in \cref{subsec:fundamental_methods}) across sequence similarity, precision, and recall. The results show that: 1) ST generally outperforms both MTSC and MTDC; 2) SFS (introduced in \cref{subsec:similar_attributes}) can enhance the performance of MTSC and MTDC, making them comparable to or even better than ST.}
\vspace{-10pt}
\label{fig:ablation-study-competition}
\end{minipage}
\hfill
\begin{minipage}[c]{0.19\textwidth}
\centering
\vspace{-5pt}
\begin{overpic}[width=\linewidth]{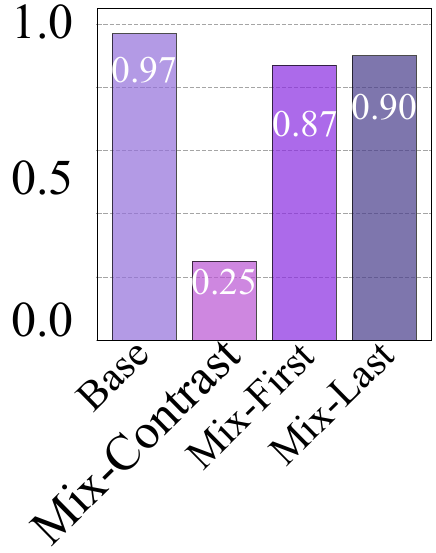}
\end{overpic}
\vspace{-14pt}
\caption{Results of the strength of SFS.}
\vspace{-10pt}
\label{fig:sfs_strength}
\end{minipage}
% \hspace{5pt}
\end{figure*}

To investigate the impact of image information usage on \task, we conduct experiments using \benchmark to compare these three fundamental methods. Specifically, to enhance robustness, we employ two different types of prompts to generate captions from GPT-4o and average the results; further details are provided in \cref{subsec:app_two_type_prompt}.
As illustrated in \cref{fig:ablation-study-competition}, ST achieves significantly higher sequence similarity, precision, and recall compared to MTSC and MTDC, indicating that greater utilization of visual information leads to improved \task performance.

% \begin{figure}[t]
% \centering
% \begin{overpic}[width=0.7\linewidth]{images/SFS.pdf}
% \put(0,-3){\small (a) Sequence similarity}
% \put(42,-3){\small (b) Precision}
% \put(76,-3){\small (c) Recall}
% \end{overpic}
% \vspace{2pt}
% \caption{
% A  comparison of ST, MTSC, and MTDC (introduced in \cref{subsec:fundamental_methods}) across \task metrics, including sequence similarity, precision, and recall. The results show that: 1) ST generally outperforms both MTSC and MTDC; 2) SFS (introduced in \cref{subsec:similar_attributes}) can enhance the performance of MTSC and MTDC, making them comparable to or even better than ST.
% }
% \vspace{-10pt}
% \label{fig:ablation-study-competition}
% \end{figure}

\subsection{Increasing the quantity of information of individual descriptions}\label{subsec:similar_attributes}

Although we have demonstrated that ST can better leverage the visual information in images and achieve strong performance on \task, directly annotating all frames of very long videos within a single dialogue may significantly degrade caption quality. Therefore, it becomes necessary to adopt MTSC or MTDC, which split frames into windows and propagate information across these windows.
Moreover, as shown in \cref{fig:ablation-study-competition}, MTSC offers no clear advantage over MTDC while incurring substantially higher computational costs due to the long history context. \textbf{Hence, it is crucial to explore how to enhance \task performance with MTDC, which solely utilizes textual information.}
While relying solely on textual information inevitably leads to some loss of detail compared to visual information, our extensive experimental exploration reveals that increasing the amount of information contained in individual descriptions enriches the conveyed content and significantly improves \task performance—achieving results comparable to, or even surpassing, those of ST.
It is important to clarify that this finding does not imply that textual information is inherently superior to visual information for \task. Rather, current LVLMs may not be specifically trained on \task-related data, which limits their ability to fully exploit visual cues. We do not address potential improvements to ST itself in this paper and leave that for future work.
Next, we introduce the process of our exploration in detail. 

\noindent\textbf{Settings of exploration:}
The core idea of this exploration is to vay the quantity of information in individual descriptions and observe its impact on \task. Specifically, we control the information quantity by varying the number of features used to describe individuals. To do this, we first collect 36 common textual features typically used to describe a person, each with several possible values (see \cref{subsec:app_list_features} for details). Then, we propose the concept of \textbf{scene format} to significantly expand the data. To create a scene format, we first select several frames from a clip in \benchmark, ensuring that the same person appears in both the first and last frames. Next, we caption each frame but remove individual descriptions from the captions, leaving blanks to form the scene format. By filling these blanks with feature values, we can flexibly control the descriptions and vary the quantity of information by adjusting the number of inserted features, then observe whether LVLMs can identify that the people in the first and last frames are the same.
\textbf{This exploration involves three modes: base, contrast, and mixture.} In base mode, individuals in the first and last frames share identical features and values. In contrast mode, shared features have different values. In mixture mode, besides shared features with the same values, an additional set of features is added only to the first frame (mixture-first) or the last frame (mixture-last).

% \textbf{1) Dataset:} We randomly select 50 video segments from MovieChat, each containing multiple frames with a person appearing in both the first and last frames. \textbf{2) Scene format:} To flexibly control the descriptions of individuals and vary their similarity, we remove the original descriptions from the captions of the selected segments, leaving blank spaces. This results in \textbf{scene formats}, which allow for flexible insertion of different descriptions.
% % Note that this experiment is conducted in the MTDC settings, with no images involved in the final process. This helps eliminate image interference and allows us to observe the impact of the descriptions.
% \textbf{3)Features' set:} We use features to create descriptions of individuals. To do this, we collect 36 common textual features that are typically used to describe a person, with each feature having several associated values. More details can be found in \cref{subsec:app_list_features}. \textbf{4) Exploration's mode:} We conduct three types of experimental modes: \textbf{base}, \textbf{contrast}, and \textbf{mixture}. In the base mode, the individuals in the first and last frames share the same set of features and values. In contrast mode, their shared features have different values. In the mixture mode, except for the shared features with the same values, we add an additional set of features only to the first frame (\textbf{mixture-first}) or the last frame (\textbf{mixture-last}).

\begin{figure*}[t]
\begin{minipage}[c]{0.44\textwidth}
\centering
\begin{overpic}[width=\linewidth]{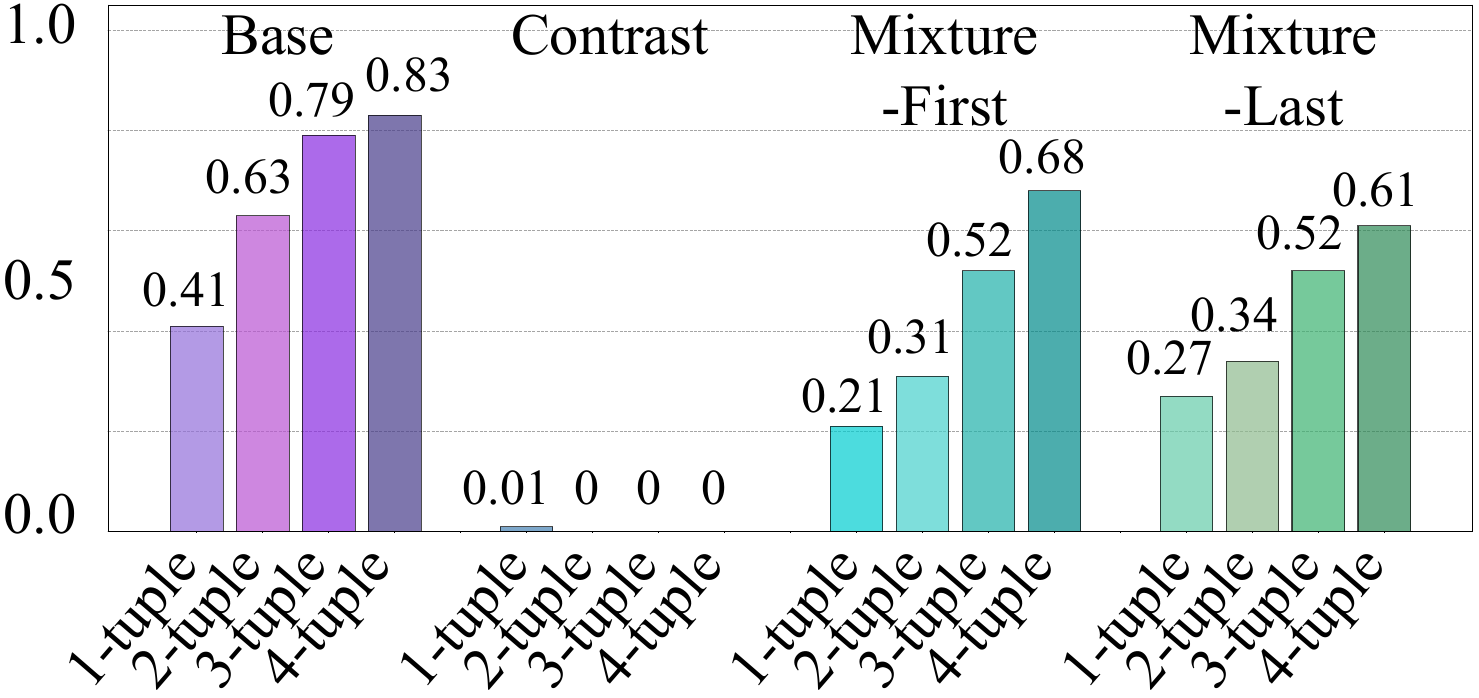}
\end{overpic}
\vspace{-12pt}
\caption{
Illustration of GPT-4o’s success rate in identifying two people as the same based on descriptions with 
$n$ features (``$n$-tuple''), with base, contrast, and mixture modes defined in \cref{subsec:similar_attributes}. The success rate and robustness improve as $n$ increases and descriptions become more similar.
}
\vspace{-5pt}
\label{fig:increase-features}
\end{minipage}
\hfill
\begin{minipage}[c]{0.48\textwidth}
\centering
\begin{overpic}[width=\linewidth]{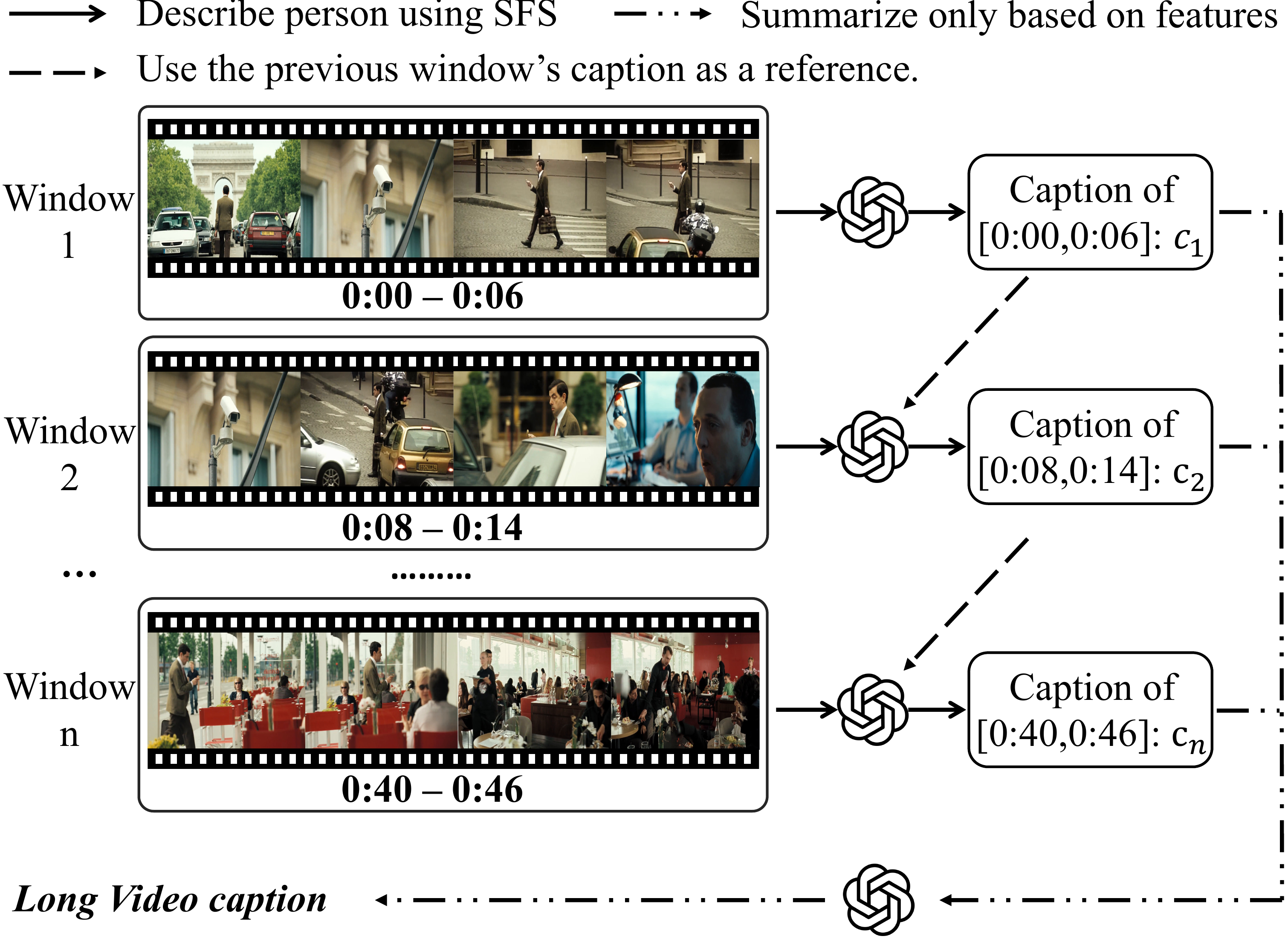}
\end{overpic}
\vspace{-10pt}
\caption{Overview of \methodabbr, consisting of two techniques, MF and ETF, detailed in \cref{subsec:our_method}.}
\vspace{-5pt}
\label{fig:rice-method}
\end{minipage}
% \hspace{5pt}
\end{figure*}

\noindent\textbf{Exploration Process:}
We gradually increase the number of features $n$ from one to four and present all experimental results in \cref{fig:increase-features}. For each $n$, we examine all possible combinations $C_m^n$, where $m$ denotes the total number of features, starting with $m=36$ when 
$n=1$. For each combination, we randomly select 5 scene formats. The result for each $n$ is obtained by averaging over all combinations and scene formats. As $n$ increases, we reduce $m$ by removing features with the lowest scores in the base mode or those whose mixture scores are significantly lower than their base scores, indicating that these features are either weak in performing \task or lack robustness and are easily influenced by others. \textbf{Results:}
As shown in \cref{fig:increase-features}, increasing $n$ improves the likelihood that GPT-4o correctly identifies the person in the first and last frames as the same individual. Furthermore, the mixture mode scores suggest that the process becomes more robust, as it is less susceptible to interference from additional features.

% \begin{figure}[t]
% \centering
% \begin{overpic}[width=0.6\linewidth]{images/mtdc_accuracy.pdf}
% \end{overpic}
% \vspace{-12pt}
% \caption{
% Illustration of the success rate of GPT-4o in identifying two people as the same when they share descriptions with $n$ features, represented in ``$n$-tuple''. (base, contrast, and mixture modes are introduced in \cref{subsec:similar_attributes}.) As 
% $n$ increases and the descriptions become more similar, the success rate grows, and the identification process becomes more robust, as it is less influenced by interference from additional features (mixture mode).
% }
% % \vspace{-16pt}
% \label{fig:increase-features}
% \end{figure}

% \begin{figure}[t]
% \centering
% \includegraphics[width=\linewidth]{images/method.pdf}
% \vspace{-15pt}
% \caption{An illustration of \methodabbr, consisting of two techniques, MF and ETF, detailed in \cref{subsec:rice}}
% \label{fig:rice-method}
% \vspace{-10pt}
% \end{figure}

\noindent\textbf{Strong feature set:}
After conducting experiments with $n=4$, we retain 11 features to form the strong feature set (SFS). We evaluate the strength of SFS by testing it in three modes: base, mixture, and \textbf{mixture-contrast}. Specifically, in the mixture-contrast mode, two individuals share identical SFS feature values but differ in another set of 4–8 features.
As shown in \cref{fig:sfs_strength}, SFS achieves a base accuracy of 97\% and a mixture accuracy of about 90\%. Interestingly, in mixture-contrast mode, even when multiple conflicting features exist, GPT-4o identifies individuals as the same person with a 25\% probability as long as their SFS feature values are identical. Furthermore, as shown in \cref{fig:ablation-study-competition}, describing individuals using all features in SFS significantly improves the performance of both MTSC and MTDC.
\textbf{SFS demonstrates strong generalizability} for the following four reasons:
1) Although we identify SFS using short clips (scene format with only few frames), SFS also works effectively on longer clips in the \benchmark dataset (each contains around 30 frames), as detailed in \cref{subsec:exp_video_caption,subsec:exp_ablation_study}.
2) We apply the same method for identifying SFS to Qwen-VL 2.5~\citep{Qwen-VL}, and find an SFS containing 12 features, 9 of which overlap with the SFS found on GPT-4o.
3) The SFS found on Qwen-VL 2.5 achieves strong performance, as detailed in \cref{subsec:exp_qwen}.
4) The SFS found on GPT-4o, when directly applied to Deepseek v3~\citep{liu2024deepseek}, achieves good results, as shown in \cref{subsec:exp_qwen}.

\begin{figure*}[t]
    \centering
    \includegraphics[width=\textwidth]{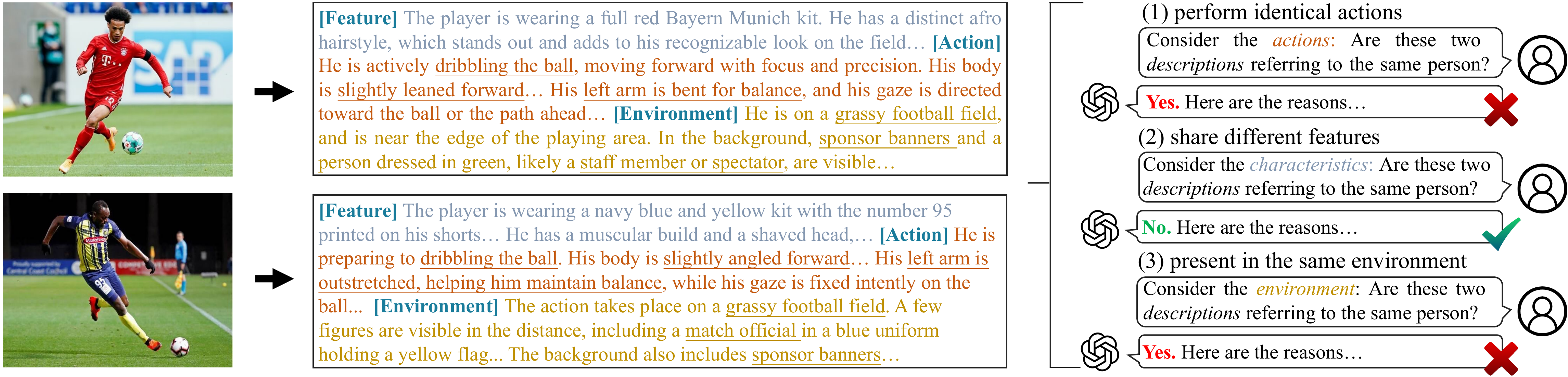}
    \vspace{-15pt}
    \caption{An example in which GPT-4o misidentifies individuals due to similar actions or similar environments. In contrast, identifying individuals based on features is more robust.}
    \vspace{-10pt}
    \label{fig:GPT-same-person-match}
\end{figure*}

\begin{figure*}[t]
% \hspace{2pt}
\begin{minipage}[c]{0.44\textwidth}
\centering
\begin{overpic}[width=\linewidth]{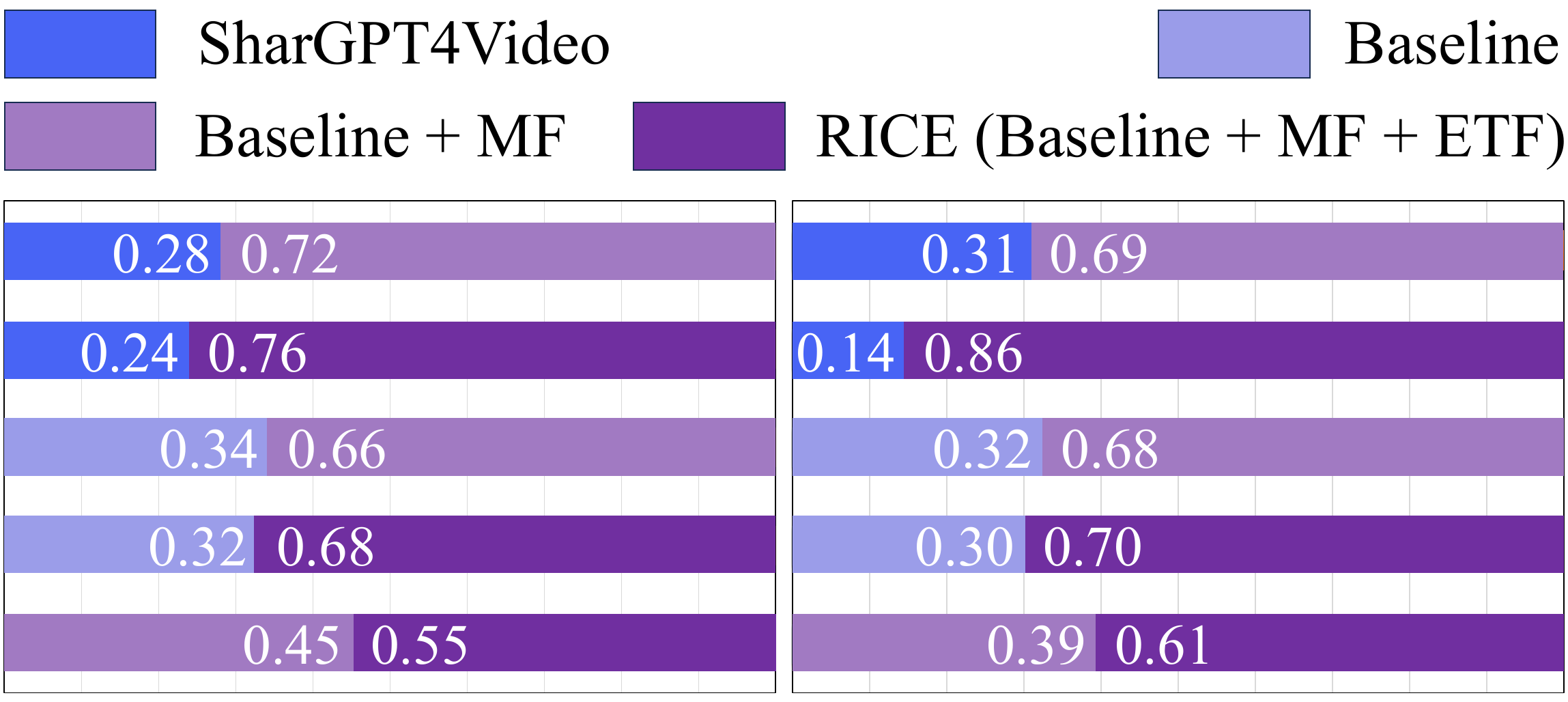}
\put(8,-8){\small (a) Activitynet}
\put(58,-8){\small (b) MovieChat}
\end{overpic}
\vspace{5pt}
\caption{\textbf{Pairwise comparisons in win rate} on two datasets.
}
\vspace{-10pt}
\label{fig:user_study}
\end{minipage}
\hfill
\begin{minipage}[c]{0.48\textwidth}
\centering
\begin{overpic}[width=\linewidth]{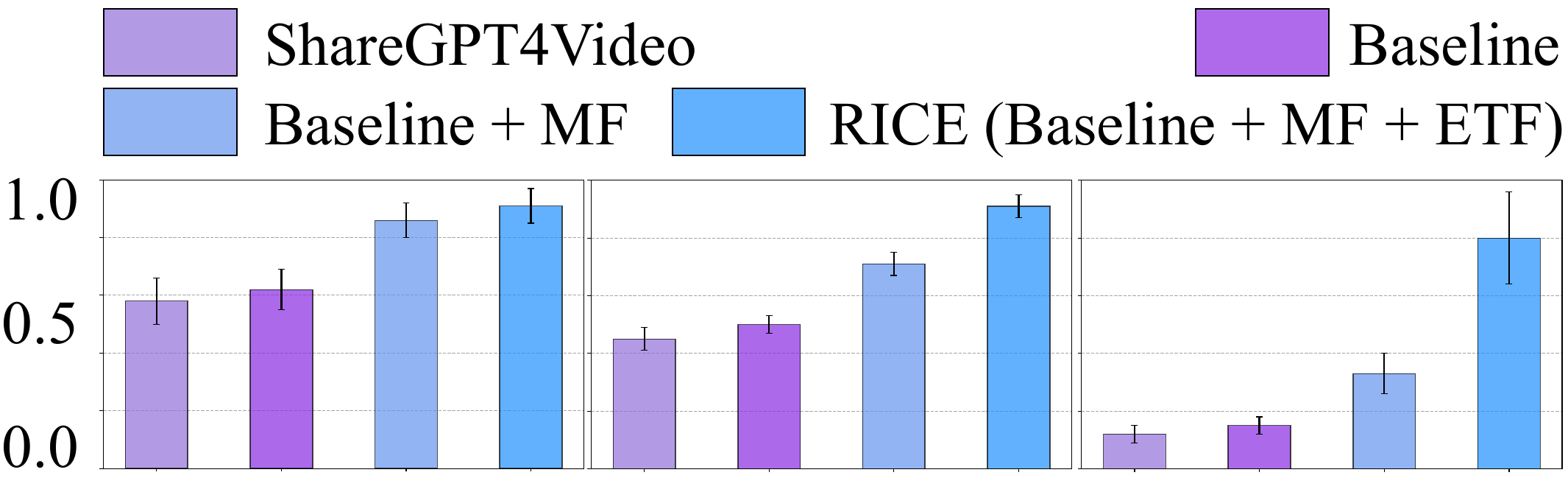}
\put(10,-4){\small (a) Sequence}
\put(14,-9){\small similarity}
\put(40,-6.5){\small (b) Precision}
\put(75,-6.5){\small (c) Recall}
\end{overpic}
\vspace{5pt}
\caption{\textbf{Comparison of \task performance} between ShareGPT4Video and \methodabbr, along with ablation studies to evaluate the effectiveness of each component of \methodabbr.}
\vspace{-10pt}
\label{fig:exp-id-matching}
\end{minipage}
% % \hspace{5pt}
\end{figure*}

\subsection{\Method}\label{subsec:our_method}
Building on the analyses presented in \cref{subsec:distance,subsec:similar_attributes}, we propose a novel video captioning method called \Method (\methodabbr), designed for implementation on LVLMs to enhance the performance of \task. As depicted in \cref{fig:rice-method}, \methodabbr comprises two key components:
\textbf{1) Multi-Frame windows (MF):} We leverage ST by inputting multiple images within the same dialogue, thereby strengthening their mutual attention. In contrast, traditional state-of-the-art methods such as ShareGPT4Video~\citep{chen2024sharegpt4video} focus primarily on differences between two adjacent frames and support input of only up to two frames at a time.
\textbf{2) Enhanced Textual Features (ETF):} We prompt GPT-4o to describe individuals and summarize across different windows based solely on the features contained in SFS.

Besides, we find that similarities in actions and environments within descriptions of individuals often lead to misidentifications, since it is common for different people to perform identical actions or appear in the same setting. An illustrative example is provided in \cref{fig:GPT-same-person-match}, with quantitative validation detailed in \cref{subsec:app_only_feature}. Fortunately, this issue can be mitigated through careful prompt design that guides LVLMs to focus exclusively on individual features, disregarding environmental and action-related cues. Therefore, we need to make sure the summarization of ETF in \methodabbr is only based on features. 
\section{Experiments}\label{sec:experiments}

% \begin{figure}[t]
%     \centering
%     \begin{overpic}[width=0.46\linewidth]{images/ablation_study.pdf}
%     \put(14,-4){\small (a) Activitynet}
%     \put(64,-4){\small (b) MovieChat}
%     \end{overpic}
%     \vspace{-10pt}
%     \caption{\textbf{Pairwise comparisons in win rate} on two datasets.}
%     \label{fig:user_study}
% \end{figure}

% \begin{figure}[t]
% \centering
% \begin{overpic}[width=0.46\linewidth]{images/exp_id_matching.pdf}
% \put(3,-4){\small (a) Sequence similarity}
% \put(42,-4){\small (b) Precision}
% \put(75,-4){\small (c) Recall}
% \end{overpic}
% \vspace{-10pt}
% \caption{
% \textbf{Quantitative comparison of \task performance} between ShareGPT4Video and \methodabbr, along with ablation studies to evaluate the effectiveness of each component of \methodabbr.
% }
% \vspace{-12pt}
% \label{fig:exp-id-matching}
% \end{figure}

\begin{figure*}[t]
    \centering
    \includegraphics[width=0.95\textwidth]{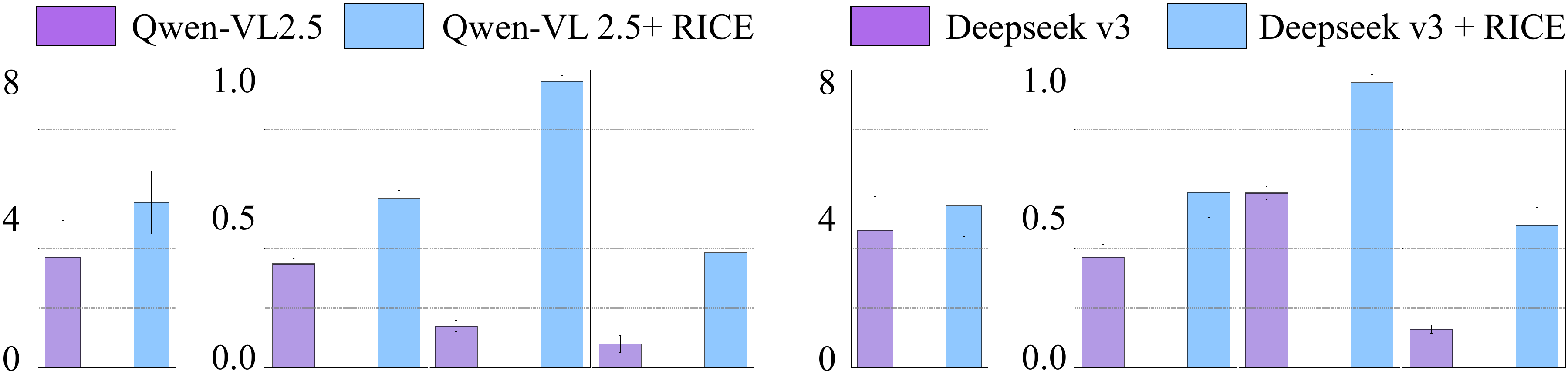}
    \vspace{-5pt}
    \caption{\textbf{Quantitative comparison} between \methodabbr implemented on Qwen-VL 2.5 and Deepseek v3 and their baseline.}
    \vspace{-10pt}
    \label{fig:exp-qwen}
\end{figure*}

\begin{table*}[t]
% \hspace{2pt}
\begin{minipage}[c]{0.38\textwidth}
\centering
\caption{
\textbf{Comparison} between SFS and the full feature set. SS represents sequential similarity. Results show that SFS outperforms full feature set on both GPT-4o and Qwen-VL 2.5. 
}
\vspace{-5pt}
\small
\label{tab:sfs}
\renewcommand{\arraystretch}{0.7}
\resizebox{\linewidth}{!}{
    \begin{tabular}{c|cc}
    \toprule
    LVLMs & Qwen-VL 2.5 & GPT-4o  \\
    \midrule
    SS (Fullset) & $0.536$ & $0.569$ \\
    SS (SFS) & $\bf{0.571}$ & $\bf{0.908}$ \\
    \midrule
    Precision (Fullset) & $0.471$ & $0.750$ \\
    Precision (SFS) & $\bf{0.954}$ & $\bf{0.910}$ \\
    \midrule
    Recall (Fullset) & $0.151$ & $0.340$ \\
    Recall (SFS) & $\bf{0.396}$ & $\bf{0.798}$ \\
    \bottomrule
\end{tabular}

}
\vspace{-10pt}
\end{minipage}
\hfill
\begin{minipage}[c]{0.58\textwidth}
\centering
% \vspace{-5pt}
\caption{
\textbf{Quantitative comparison of caption quality} between ShareGPT4Video and \methodabbr, along with ablation studies to evaluate the effectiveness of each component of \methodabbr. Note that Text-Coverage cannot be calculated on \benchmark.
}
\vspace{10pt}
\label{tab:ablation-study}
\renewcommand{\arraystretch}{0.8}
\resizebox{\linewidth}{!}{
    \begin{tabular}{c|cc|c}
    \toprule
    \multirow{2}{*}{Method} & \multicolumn{2}{c|}{\textbf{ActivityNet}} & \multicolumn{1}{c}{\textbf{\benchmark}}  \\
     &\textbf{GPT-score}$\uparrow$ & \textbf{Text-Coverage}$\uparrow$  &\textbf{GPT-score}$\uparrow$\\
    \midrule
    ShareGPT4Video & $4.21\pm2.20$ & $0.59\pm0.32$ & $3.01\pm1.67$ \\
    Baseline & $4.43\pm1.81$ & $0.61\pm0.27$ & $3.34\pm1.81$ \\
    Baseline+MF & $5.71\pm1.31$ & $0.67\pm0.13$ & $4.88\pm1.92$ \\
    \methodabbr(MF+ETF) & $\bf{6.04\pm1.21}$ & $\bf{0.69\pm0.18}$ & $\bf{5.91\pm1.91}$  \\
    \bottomrule
\end{tabular}

}
\vspace{-10pt}
\label{fig:quantitative_comparison}
\end{minipage}
% \hspace{5pt}
\end{table*}

% \begin{figure*}[t]
%     \centering
%     \includegraphics[width=\textwidth]{images/video_tracking.pdf}
%     \vspace{-20pt}
%     \caption{An example of tracking in a long video with multiple transitions, where each person is shown in a different color.}
%     \vspace{-5pt}
%     \label{fig:tracking}
% \end{figure*}

In this section, we conduct extensive experiments to evaluate \methodabbr. First, \cref{subsec:exp_setting} details the experimental settings. Then, \cref{subsec:exp_video_caption} compares \methodabbr with the SOTA method ShareGPT4Video~\citep{chen2024sharegpt4video} using four metrics. Next, \cref{subsec:exp_ablation_study} presents ablation studies to assess each component’s contribution. After that, we apply \methodabbr to other LVLMs—Qwen-VL 2.5~\citep{Qwen-VL}, and Deepseek v3~\citep{liu2024deepseek}—and compare them with their baselines in \cref{subsec:exp_qwen}. 
% Finally, we leverage GPT-4o’s \task capabilities to track long videos with shot changes, providing enhanced location information for captions.

\subsection{Experimental settings}\label{subsec:exp_setting}
For experiments in \cref{subsec:exp_video_caption,subsec:exp_ablation_study}, we use ActivityNet~\citep{caba2015activitynet} and \benchmark. For all ShareGPT4Video experiments in these sections, we use GPT-4o instead of GPT-4V (as in their paper) for a fair comparison. Additionally, we set the MF length to 4. The difference between ShareGPT4Video and the baseline (in \cref{fig:user_study,fig:exp-id-matching,fig:exp-qwen}) lies in the used prompts: the baseline uses our designed prompt without accounting for differences between adjacent frames. Except for experiments in \cref{subsec:exp_qwen}, all experiments in this section are conducted on GPT-4o.

\noindent\textbf{Evaluation metrics.}
Traditional metrics such as BLEU~\citep{papineni2002bleu} and METEOR~\citep{denkowski2014meteor} are inadequate for evaluating highly detailed captions, particularly when detailed annotations are lacking. Instead, we use four evaluation metrics, including: \textbf{1) User preference study}: We conduct a user preference study with 10 annotators selecting which of two candidates they prefer, detailed in \cref{subsec:app_user_study}. \textbf{2) GPT-score}: We use GPT-4o as an evaluator, scoring based on the ground truth annotated captions following the methodologies outlined in~\citet{zheng2023judging,wang2023large}, detailed in \cref{subsec:app_gpt_score}. \textbf{3) Text-coverage}: For datasets with annotations for different segments, like ActivityNet, we introduce a new metric called Text-coverage. Given the ground truth set $g=\{g_1, g_2, \dots, g_k\}$ (where $k$ is the number of segments for this video), we calculate the Text-coverage score $TC(g, p)$ of prediction $p$ as $TC(g, p) = \sum_{i=1}^k s(g_i, p)$. Here, $s(g_i, p)$ is defined as
\def\T5{\mathrm{T5}}
% \vspace{-8pt}
\begin{equation}
s(g_i, p) =
\begin{cases}
1 & \text{if} \displaystyle\max\limits_{1 \le j \le n_j}\frac{\langle \T5(g_i), \T5(p_j) \rangle}{\Vert \T5(g_i) \Vert_2 \Vert \T5(p_j) \Vert_2} \ge \rho \\
0 & \text{otherwise}
\end{cases}
% \vspace{-5pt}
\end{equation}
In this equation, $p_j$ is the $j$-th sub-sentence of the predicted video caption, and $\rho$ is a threshold. \textit{A higher $TC(g, p)$ indicates that more segments from ground truth match with similar versions in the predicted video caption, indicating a better video caption overall.} \textbf{4) \task score}: For the videos in \benchmark, we use the metrics mentioned in \cref{subsubsec:eval_metrics} to evaluate the performance of \task.

\subsection{Evaluation on quality of video captions}\label{subsec:exp_video_caption}

We compare \methodabbr with the state-of-the-art method, ShareGPT4Video~\citep{chen2024sharegpt4video}, using the settings and metrics introduced in \cref{subsec:exp_setting}.
\textbf{Results: } 
\textbf{1) User Preference Study}: As shown in \cref{fig:user_study}, \methodabbr consistently receives a higher user preference score than ShareGPT4Video (the second row). \textbf{2) GPT-score and Text-coverage}: As shown in \cref{fig:quantitative_comparison}, \methodabbr achieves better GPT-score and Text-coverage score than ShareGPT4Video on evaluated datasets. \textbf{3) \task Score}: As illustrated in \cref{fig:exp-id-matching}, \methodabbr achieves higher \task scores in terms of sequence similarity, precision, and recall on \benchmark. Overall, \methodabbr outperforms ShareGPT4Video.

\subsection{Ablation studies}\label{subsec:exp_ablation_study}
\noindent\textbf{Settings.}
To evaluate the contributions of the two \methodabbr techniques, MF and ETF, we compare \methodabbr (baseline + MF + ETF) with the baseline and baseline + MF, using the datasets and metrics from \cref{subsec:exp_setting}. We also compare descriptions generated using SFS versus the full feature set, as shown in \cref{tab:sfs}. \textbf{Results.}
As shown in \cref{fig:user_study} (User preference study), \cref{tab:ablation-study} (GPT-score and Text-coverage), and \cref{fig:exp-id-matching} (\task performance), both MF and ETF significantly improve performance. Moreover, \cref{tab:sfs} shows that SFS outperforms the full feature set.

\subsection{Implementation on LVLMs except for GPT-4o}\label{subsec:exp_qwen}
Besides GPT-4o, we apply \methodabbr to, Qwen-VL 2.5~\citep{Qwen-VL}, and Deepseek v3~\citep{liu2024deepseek}, comparing against their baselines using GPT-score and \task metrics from \cref{subsec:exp_setting}. As shown in \cref{fig:exp-qwen}, \methodabbr consistently outperforms the baselines across all metrics and models. Notably, the SFS for Qwen-VL and Deepseek v3 is derived from GPT-4o, while Qwen-VL 2.5 uses its own SFS, demonstrating SFS’s generalizability across models.

% \subsection{Multi-modal understanding}\label{subsec:exp_mm_understanding}
% \todo{
% We compare our \methodabbr with ShareGPT4Video by comparing the performance of \vlm and ShareGPT4Video-8B on widely used multimodal benchmarks MVBench\citep{videollava-7b},
% as the only difference between \vlm and ShareGPT4Video-8B are the source of training data is from \methodabbr or ShareGPT4Video. 
% As shown in \cref{tab:vqa_mvbench}, \vlm outperforms ShareGPT4Video-8B, demonstrating that \methodabbr is superior to ShareGPT4Video.}

% \subsection{Obatining location information using long video tracking}\label{subsec:exp_video_tracking}
% Inspired by image captioning methods that provide location information~\citep{yang2024bacon}, we propose using video tracking methods to enhance video captions with location information. However, traditional tracking methods, such as SAM2~\citep{ravi2024sam}, struggle to track long videos that contain shot changes because they rely on the continuous change of segmentation masks, while the connectivity is lost during shot changes.
% Fortunately, we can leverage the \task capabilities of GPT-4o in conjunction with SAM2 to enable long video tracking that accommodates transitions. 
% Detailed methods and more examples can be found in \cref{subsec:app_tracking}. A simple example of tracking across transitions in ~\cref{fig:tracking}.
\section{Conclusion}\label{sec:conclusion}

In this paper, we explore the \task problem in long video captioning by introducing \benchmark, a comprehensive toolset including a dataset, an automatic prediction extractor, and three evaluation metrics. Leveraging this benchmark, we analyze GPT-4o and identify two key insights to improve \task performance: 1) enhancing the usage of visual information and 2) increasing the quantities of information of individual descriptions. Based on them, we propose \methodabbr, a method for continuous tracking of individuals in long video captions. Extensive experiments validate the effectiveness of our approach. We hope our work offers valuable insights and resources to advance video captioning research.

\noindent\textbf{Limitations.}
One of the most significant limitations is that the datasets and benchmarks we constructed are limited in scale due to resource constraints.
Furthermore, this paper does not address extremely challenging scenarios, such as when a person's clothing changes and facial features are insufficient. Our work focuses on improving performance in cases where sufficient features are available for \task. Finally, we concentrated on character recognition. While we believe similar phenomena may apply to ordinary objects, we did not explore these extensions due to resource limitations. We leave this for future work.

% \begin{ack}
% Use unnumbered first level headings for the acknowledgments. All acknowledgments

% Do {\bf not} include this section in the anonymized submission, only in the final paper. You can use the \texttt{ack} environment provided in the style file to automatically hide this section in the anonymized submission.
% \end{ack}
{
% \small
\bibliographystyle{abbrvnat}
\bibliography{ref}
}
\newpage
\appendix
\newcommand{\AppendixPrefix}{A}
\setcounter{section}{0}
\renewcommand{\thefigure}{\AppendixPrefix\arabic{figure}}
\setcounter{figure}{0}
\renewcommand{\thetable}{\AppendixPrefix\arabic{table}}
\setcounter{table}{0}
\renewcommand{\theequation}{\AppendixPrefix\arabic{equation}}
\setcounter{equation}{0}

% \onecolumn
\section{Overview of the appendix}\label{sec:app_overview}
The supplementary materials consist of the following parts:
\begin{itemize}
    \item \textbf{Additional details of \benchmark}, including four parts:
    \begin{itemize}
        \item We provide complete details on data collection and present a full example. (corresponding to \cref{subsubsec:dataset} of the main paper and detailed in \cref{subsec:app_collecting_data})
        \item We provide the details of the method used to extract predictions from captions. (corresponding to \cref{subsubsec:extract_method} of the main paper and detailed in \cref{subsec:app_extract_method})
        \item We provide the details of calculating metrics given the extracted sequence of predictions. (corresponding to \cref{subsubsec:eval_metrics} of the main paper and detailed in \cref{subsec:app_calculate_metrics})    
        \item General ethical conduct for \benchmark. (detailed in \cref{subsec:app_ethical_conduct})
    \end{itemize}

    \item \textbf{Additional details of analysis}, including three parts:
    \begin{itemize}
        \item Additional details of the three methods for captioning multiple frames, ST, MTSC and MTDC. (corresponding to \cref{subsec:fundamental_methods} of the main paper and detailed in \cref{subsec:app_three_basic_methods})
        \item Two types of used prompt in the analysis part. (corresponding to \cref{subsec:distance} of the main paper and detailed in \cref{subsec:app_two_type_prompt})
        \item The complete list of the 50 features used for exploration. (corresponding to \cref{subsec:similar_attributes} of the main paper and detailed in \cref{subsec:app_list_features})
        \item The detailed quantitative experiments demonstrating the importance of removing action and environment components from the description. (corresponding to \cref{subsec:our_method} of the main paper and detailed in \cref{subsec:app_only_feature})
    \end{itemize}

    \item \textbf{Additional details of \methodabbr}, mainly including the used prompts. (corresponding to \cref{subsec:our_method} of the main paper and detailed in \cref{sec:app_rice})

    \item \textbf{Additional details of experiments}, including five parts:
    \begin{itemize}
        \item We train a \vlm using the data annotated by \methodabbr (detailed in \cref{subsec:app_training}).
        \item The details about the user study. (corresponding to \cref{subsec:exp_setting} of the main paper and detailed in \cref{subsec:app_user_study})
        \item The prompts used for calculating the GPT-score. (corresponding to \cref{subsec:exp_setting} of the main paper and detailed in \cref{subsec:app_gpt_score})
        % \item The detailed experimental results of multi-modal understanding. (corresponding to \cref{subsec:exp_mm_understanding} of the main paper and detailed in \cref{subsec:app_mmunderstanding}).
        \item We use the capabilities of LVLMs to enable long video tracking. (detailed in \cref{subsec:app_tracking}).
    \end{itemize}

\end{itemize}

\section{Additional details of \benchmark}\label{sec:app_benchmark}

\subsection{Additional details of collecting data}\label{subsec:app_collecting_data}

We annotate 374 segments from long videos in MovieChat~\citep{song2023moviechat}, Miradata~\citep{ju2024miradata} and Panda-70M~\citep{chen2024panda}, ensuring each segment containing multiple characters appear repeatedly. Next, we employ the method presented in ShareGPT4Video~\citep{chen2024sharegpt4video} to extract key frames from these video segments.

\begin{figure*}[t]
\centering
\includegraphics[width=\linewidth]{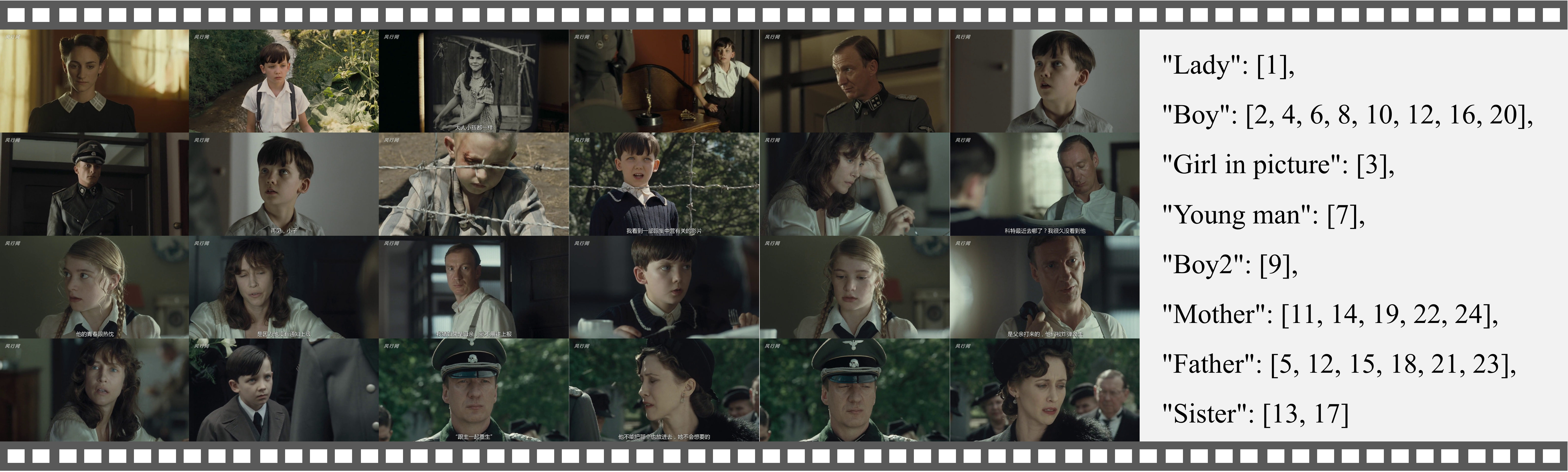}
\vspace{-15pt}
\caption{A complete example from \benchmark.}
\label{fig:app-dataset-example}
\end{figure*}

Subsequently, we conduct manual annotations with two independent groups of volunteers. To prevent fatigue and maintain annotation quality, we limit volunteers to four hours of annotation per day.
Initially, both groups perform preliminary annotations on the videos, focusing on identifying the number of characters and outlining the general storyline. After this, we divide the dataset into two equal parts, with each group responsible for providing detailed annotations for their assigned half. These detailed annotations include verifying the presence and roles of characters in each video segment.
Once the initial annotations are complete, the two groups exchange their results and conduct cross-checks. Any discrepancies are resolved by reevaluating the corresponding video segments to ensure consistency and accuracy. The final dataset is created by merging the annotations from both groups, resulting in a comprehensive and thoroughly verified video dataset. An example of the annotated data is shown in \cref{fig:app-dataset-example}.

Each segment of the result dataset has approximately 30 frames and 8.27 people, and each individual appear 2.77 times in average.
The distribution of the length of the intervals between consecutive appearances of the same person is shown in \cref{fig:data-distribution}.

\subsection{Additional details of the method to extract predictions from captions}\label{subsec:app_extract_method}

\begin{figure*}[t]
\centering
\includegraphics[width=\textwidth]{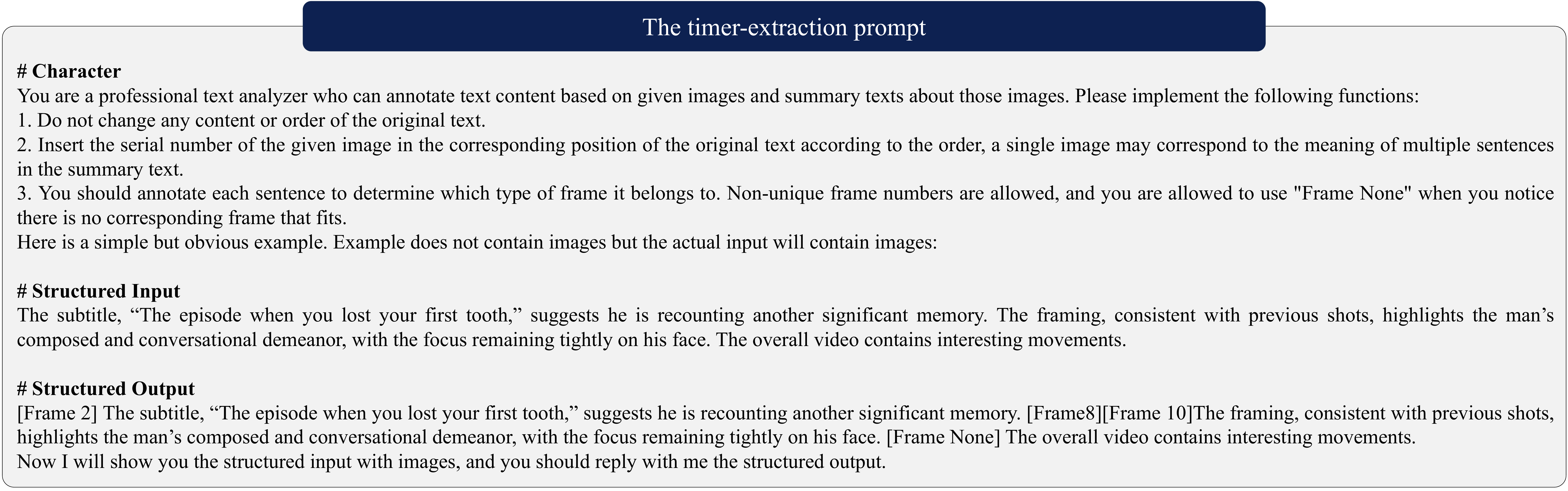}
\vspace{-10pt}
\caption{The time-stamping prompt is employed exclusively for extracting timestamps. Testing on 100 samples and validated by volunteers, GPT-4o achieved a $100\%$ match with volunteer-generated results.}
\label{fig:prompt-b2-step1}
\end{figure*}

\begin{figure*}[t]
\centering
\includegraphics[width=\textwidth]{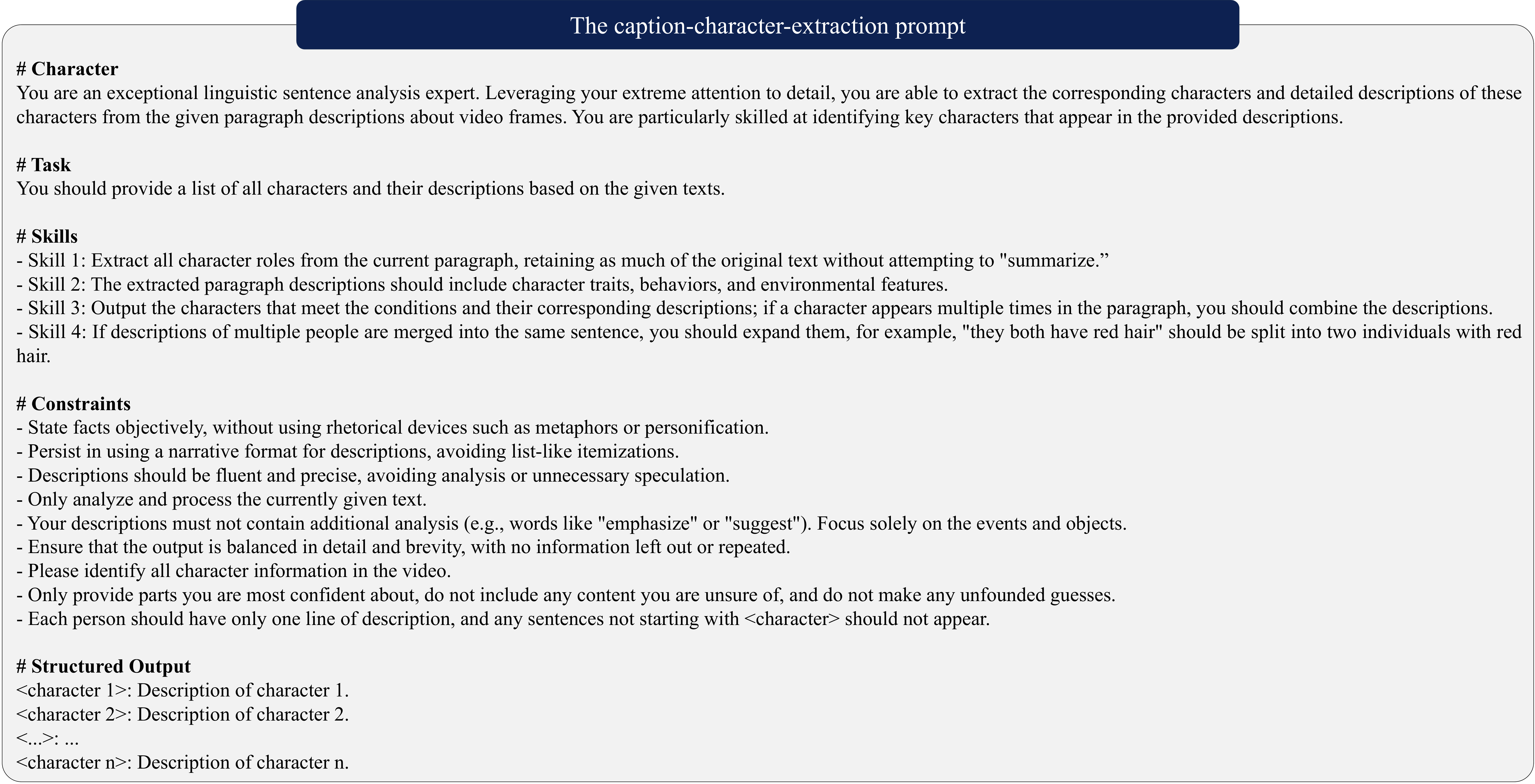}
\vspace{-10pt}
\caption{The character-extraction prompt is applied to extract character information from descriptions. When tested on 100 samples and validated by volunteers, GPT-4o achieved $97\%$ consistency with volunteer results.}
\label{fig:prompt-b2-step2}
\end{figure*}

\begin{figure*}[t]
\centering
\includegraphics[width=\textwidth]{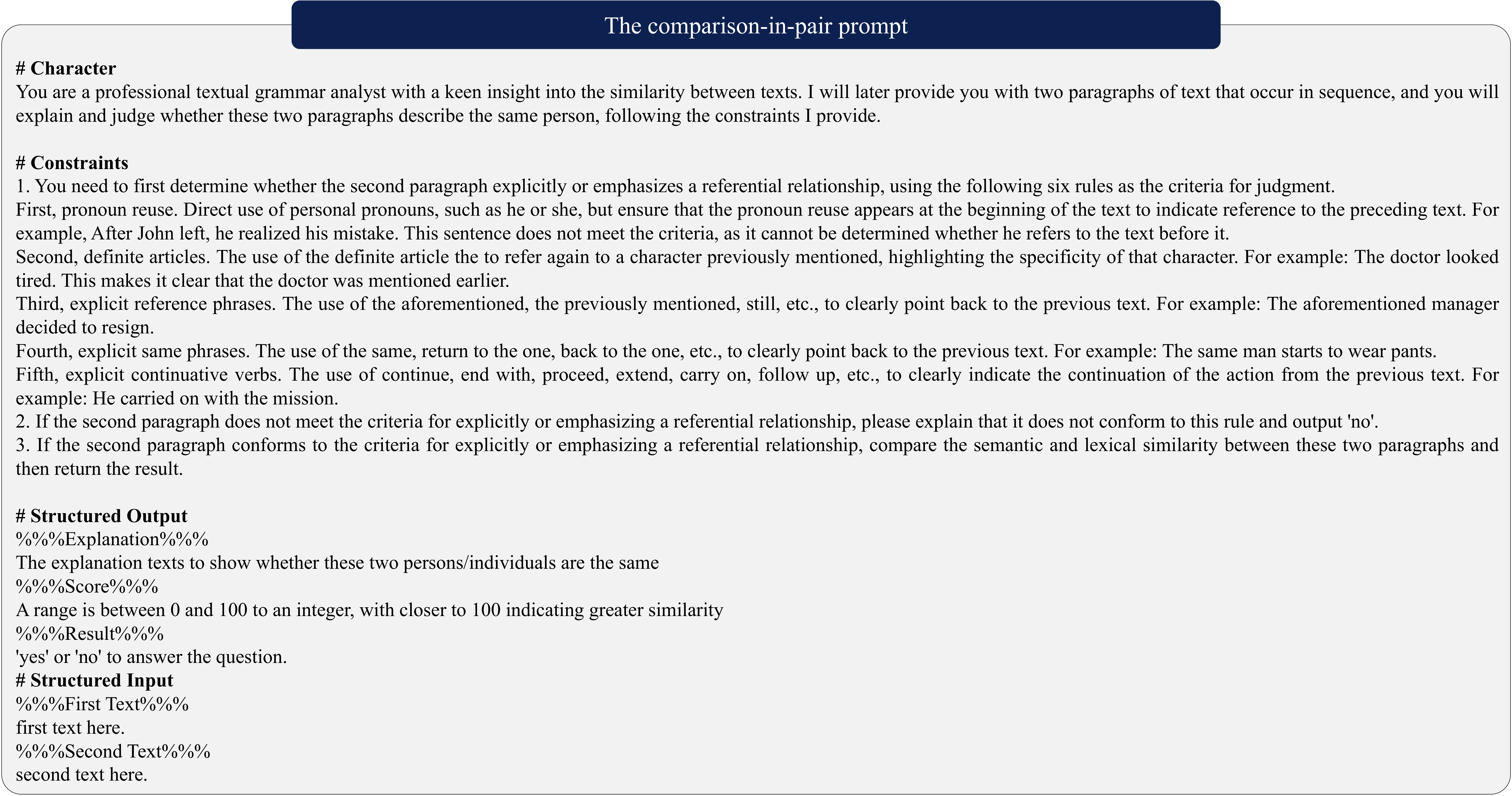}
\vspace{-10pt}
\caption{To mitigate the high cost of volunteer-based scoring, the judge prompt is applied in the scoring step. Testing on 100 samples, GPT-4o achieved $90\%$ agreement with volunteer evaluations. As the lowest agreement rate observed was $90\%$, we conclude that this approach demonstrates over $90\%$ similarity with manual extraction in the main paper, underscoring its reliability.}
\label{fig:prompt-b2-step3}
\end{figure*}

The method for extracting predictions from captions consists of three steps:

\noindent \textbf{Step 1: Assign frame indices to the video caption and split it into segments.}

To extract the predicted sequence, we first need to assign frame indices to the video captions, which typically do not have indices. We use GPT-4 to perform this assignment, and the prompt we used is shown in \cref{fig:prompt-b2-step1}. It’s important to note that we ensure no words are altered during this process by carefully designing the prompt, as phrases like “the person” are crucial for identifying individuals. Once we assign the frame indices, we can easily split the caption into multiple parts, with each part corresponding to a specific frame.

\noindent \textbf{Step2: Extract character description from the descriptions.}

In this step, we extract the descriptions of all objects that appear in each frame from the text associated with that frame. Similar to step 1, we carefully design the prompt to ensure that the original wording of each description remains unchanged. The prompt we used is shown in \cref{fig:prompt-b2-step2}.

\noindent \textbf{Step3: Obtaining the predicted \task sequence}

We implement a dynamic sequence updating mechanism to obtain the predicted \task sequence.
First, we initialize the character list with distinct characters from the first frame. Subsequently, for each new frame, we compare its characters with the existing list to find the best match. Matches exceeding a threshold are updated into the existing list; otherwise, a new character entry is created. Details are shown in \cref{alg:dynamic_sequence_update}. For the checking function, we use the prompt in \cref{fig:prompt-b2-step3}.

\begin{algorithm}
\caption{Dynamic Sequence Updating Algorithm}
\label{alg:dynamic_sequence_update}
\begin{algorithmic}[1]
\REQUIRE $\textit{frame\_description\_dict}$, function \textit{find\_most\_fit}
\ENSURE Updated character list: $\textit{character\_list}$

\STATE Initialize from the first frame
\FOR{$\textit{frame\_info}$ in $\textit{frame\_description\_dict}$}
    \FOR{$\textit{character}$ in $\textit{frame\_info}$}
        \STATE \textit{idx}, \textit{score} $\gets$ \textit{find\_most\_fit}(\textit{character\_list}, \textit{char})
        \IF{$\textit{score} > \textit{threshold}$}
            \STATE \textit{character\_list}[\textit{idx}].\textit{update}(\textit{char})
        \ELSE
            \STATE \textit{character\_list}.\textit{append}(\textit{char})
        \ENDIF
    \ENDFOR
\ENDFOR

\STATE \RETURN $\textit{character\_list}$
\end{algorithmic}
\end{algorithm}

\subsection{Additional details of calculating metrics}\label{subsec:app_calculate_metrics}

In this section, we provide the details of calculating the metrics given the predicted sequence and the ground truth.
To achieve this, we use a bipartite graph, where each node represents a character, and edges (with weights denoting sequence overlaps) connect nodes with potential mappings. The maximum flow in this graph provides a robust measure of \task quality. For example, given a ground truth sequence(GT) and the predicted sequence(PD):
\begin{verbatim}
GT: <bean>: [2, 4, 5, 7, 8, 9, 10]
\end{verbatim}
\begin{verbatim}
PD: <character 2>: [2, 4, 5, 6, 7, 8]
\end{verbatim}
matches well, as confirmed by the bipartite graph with weight $5$ for $[2,4,5,7,8]$.

Next, we calculate the three metrics as follows:

1. \textbf{Sequence similarity}: The sequence similarity is computed as the ratio of the maximum flow weight from the bipartite graph to the total ground truth characters:
\[
\text{Accuracy} = \frac{\text{WeightAns}}{\text{TotalAns}}
\]
where $\text{WeightAns}$ is the maximum flow and $\text{TotalAns}$ is the ground truth total.

2. \textbf{Recall and Precision}: These metrics evaluate pairwise matches between predicted and ground truth sequences. For a character like $<\text{character 2}>$, predicted pairs such as $(2, 4)$ and $(7, 8)$ are compared against ground truth pairs. True positives ($\text{TP}$) are defined as:
\[
\text{TP} = \sum((\text{groundTruth\_count} \land \text{prediction\_count}).\text{values()})
\]
From this, recall and precision are calculated as:
\[
\text{Recall} = \frac{\text{TP}}{\text{Total Ground-Truth-Pairs}}, \quad \quad \text{Precision} = \frac{\text{TP}}{\text{Total Prediction-Pairs}}
\]

\subsection{General ethical conduct}\label{subsec:app_ethical_conduct}
Since our work involves using human-derived data, it is essential to clarify the compliance and ethical considerations associated with the datasets used in this study.

\textbf{1) Privacy Protection.} The datasets we employ do not include any personally identifiable information (PII) or sensitive personal data. During the annotation process, character names are assigned using general descriptors like \textit{man} and \textit{woman}, effectively eliminating any risk of privacy violations.

\textbf{2) Bias Mitigation.} The datasets used in this study are sourced from well-established and publicly available benchmarks, including MovieChat~\citep{song2024moviechat} MovieChat~\citep{song2023moviechat}, Miradata~\citep{ju2024miradata} and Panda-70M~\citep{chen2024panda}. These datasets are widely recognized for their adherence to ethical standards, ensuring that their use does not introduce or exacerbate any biases.

\textbf{3) Annotator Welfare.} The annotation process is conducted with a strong focus on annotator welfare. Each annotator is limited to a maximum of 4 working hours per day, with mandatory breaks to ensure sufficient rest and prevent fatigue. This approach reflects our commitment to upholding human rights and promoting ethical working conditions.

\section{Additional details of analysis}\label{sec:app_analysis}

\subsection{Additional details of ST, MTSC and MTDC}\label{subsec:app_three_basic_methods}

In this section, we present a detailed overview of the three methods for captioning using GPT-4o: ST, MTSC, and MTDC, along with the prompts utilized for each method.

\textbf{1) ST:} In the ST method, all frames are input directly into the model in a single dialogue for captioning.

\textbf{2) MTSC:} The MTSC method involves providing different frames in separate dialogues to GPT-4o, though these frames share the same context. Additionally, there is variability in whether the caption of the previous frame is provided while captioning a new frame. If the caption of the last frame is included, we denote this as MTSC-text; conversely, if it is not provided, we refer to it as MTSC-notext. Unlike the ST approach, MTSC requires a summary stage to derive the final caption based on the captions of all frames.

\textbf{3) MTDC:} In the MTDC method, different frames are presented to GPT-4o in distinct contexts. When captioning a frame, the model does not have access to the images of preceding frames. Similar to MTSC, MTDC also necessitates a summary stage to obtain the final caption.

Specifically, we utilize the same prompt for both ST and the captioning stages of MTSC and MTDC, as illustrated in \cref{fig:prompt_st_mtsc_mtdc_caption_stage}. Additionally, the same prompt is employed during the summary stages of both MTSC and MTDC, as shown in \cref{fig:prompt_mtsc_mtdc_summary_stage}.

\begin{figure*}[t]
% % \hspace{2pt}
\begin{minipage}[c]{0.29\textwidth}
\centering
% \vspace{2pt}
\begin{overpic}[width=0.99\linewidth]{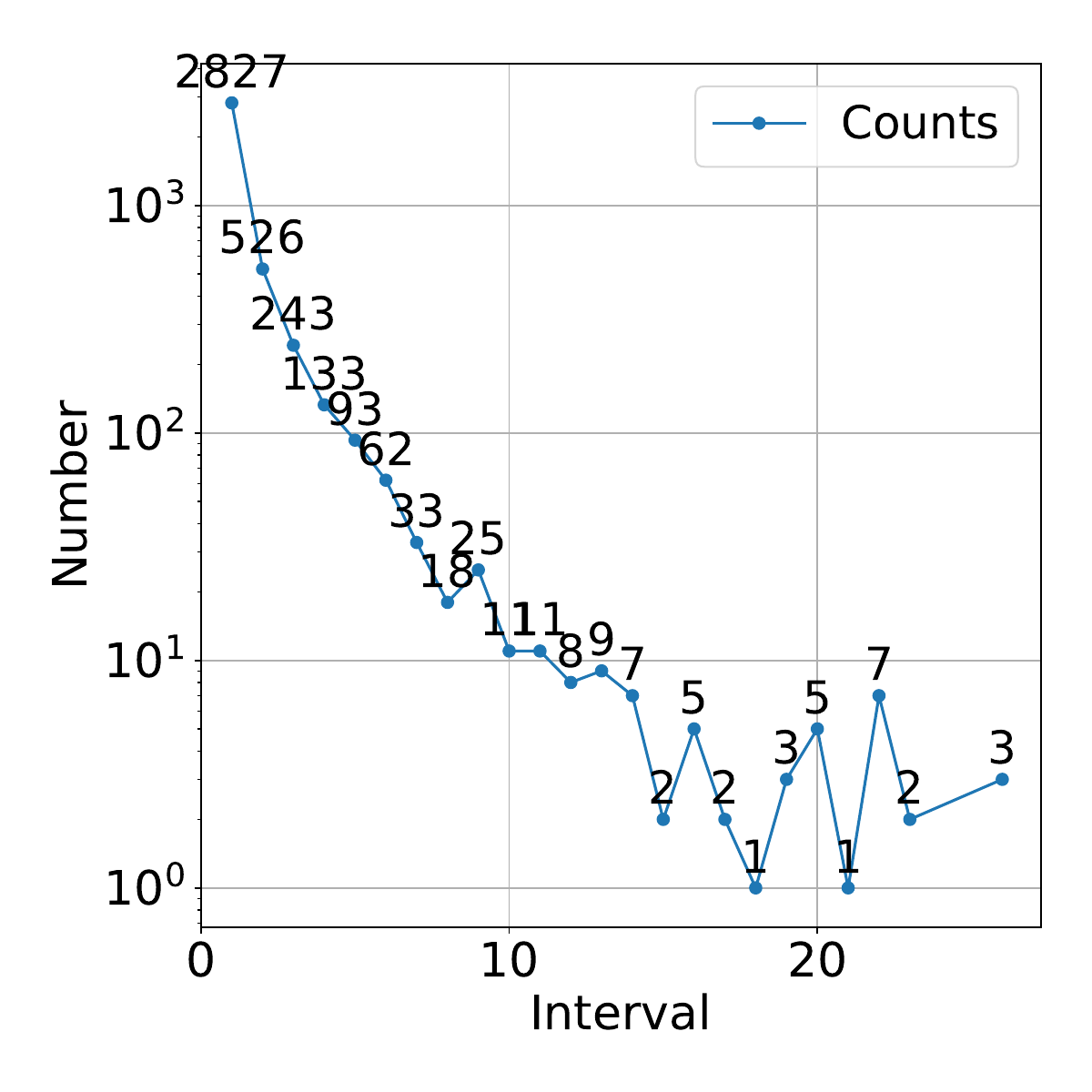}
\end{overpic}
\vspace{1pt}
\caption{
The distribution of the lengths of intervals between consecutive.
}
% \vspace{-16pt}
\label{fig:data-distribution}
\end{minipage}
\hfill
\begin{minipage}[c]{0.63\textwidth}
\centering
\begin{overpic}[width=0.99\linewidth]{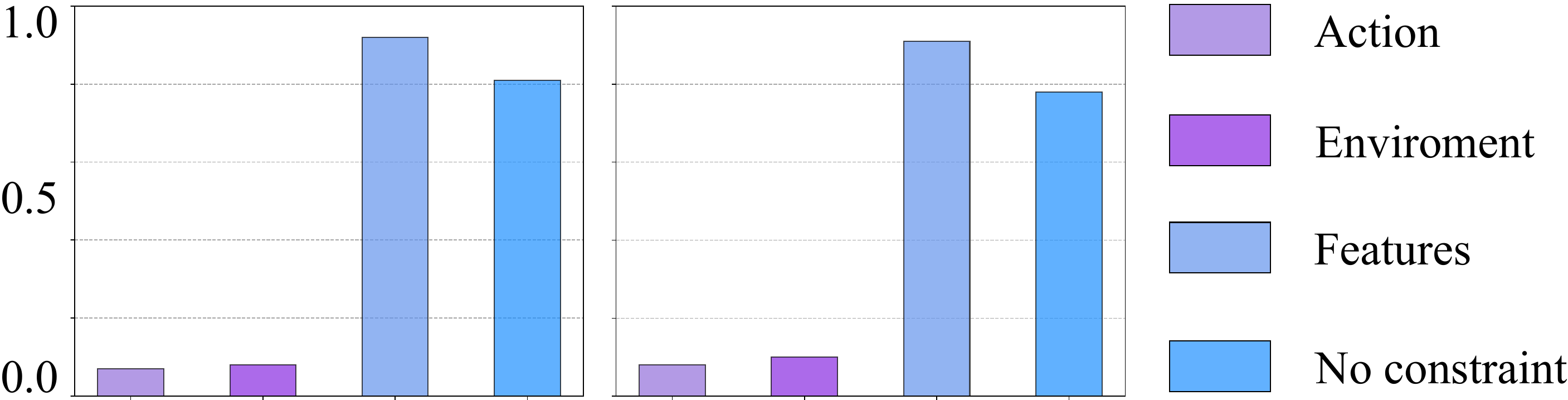}
  \put(12,-3){\small (a)Precision}
  \put(48,-3){\small (b)Recall}
\end{overpic}
\vspace{6pt}
\caption{
Quantitative comparison between criteria used to determine whether two individuals appearing in different frames of a video are the same person, including based on environment, actions, features, or all of them. The results are evaluated using precision and recall. The findings indicate that determining identity based solely on features is the most robust method compared to the others.
}
% \vspace{-10pt}
\label{fig:only-env}
\end{minipage}
% \hspace{5pt}
\end{figure*}

\subsection{Two types of used prompt in the analysis part}\label{subsec:app_two_type_prompt}

\begin{figure*}[t]
\centering
\includegraphics[width=\textwidth]{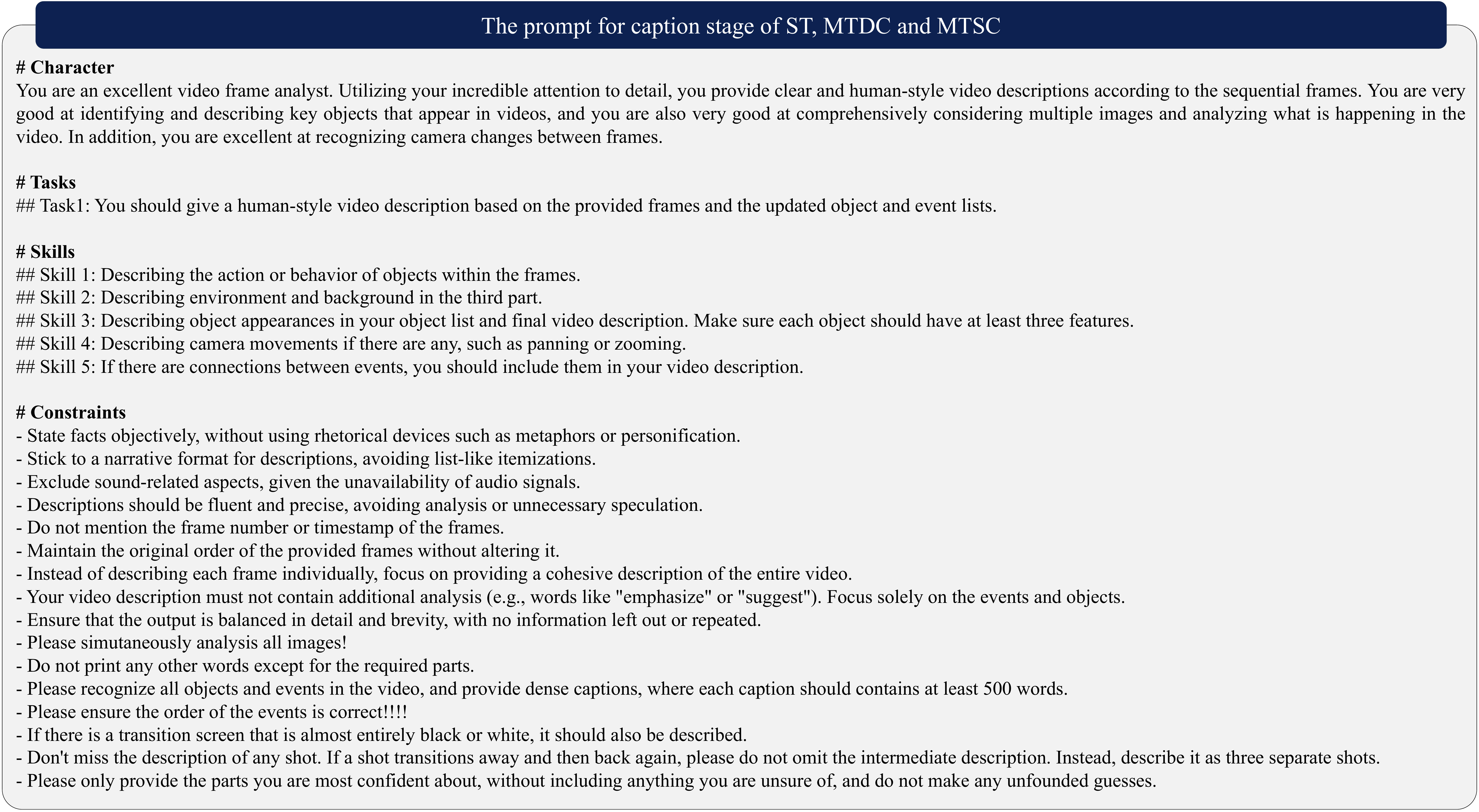}
\vspace{-10pt}
\caption{The prompt used for the caption stage of ST, MTSC, and MTDC.}
\label{fig:prompt_st_mtsc_mtdc_caption_stage}
\end{figure*}

\begin{figure*}[t]
\centering
\includegraphics[width=\textwidth]{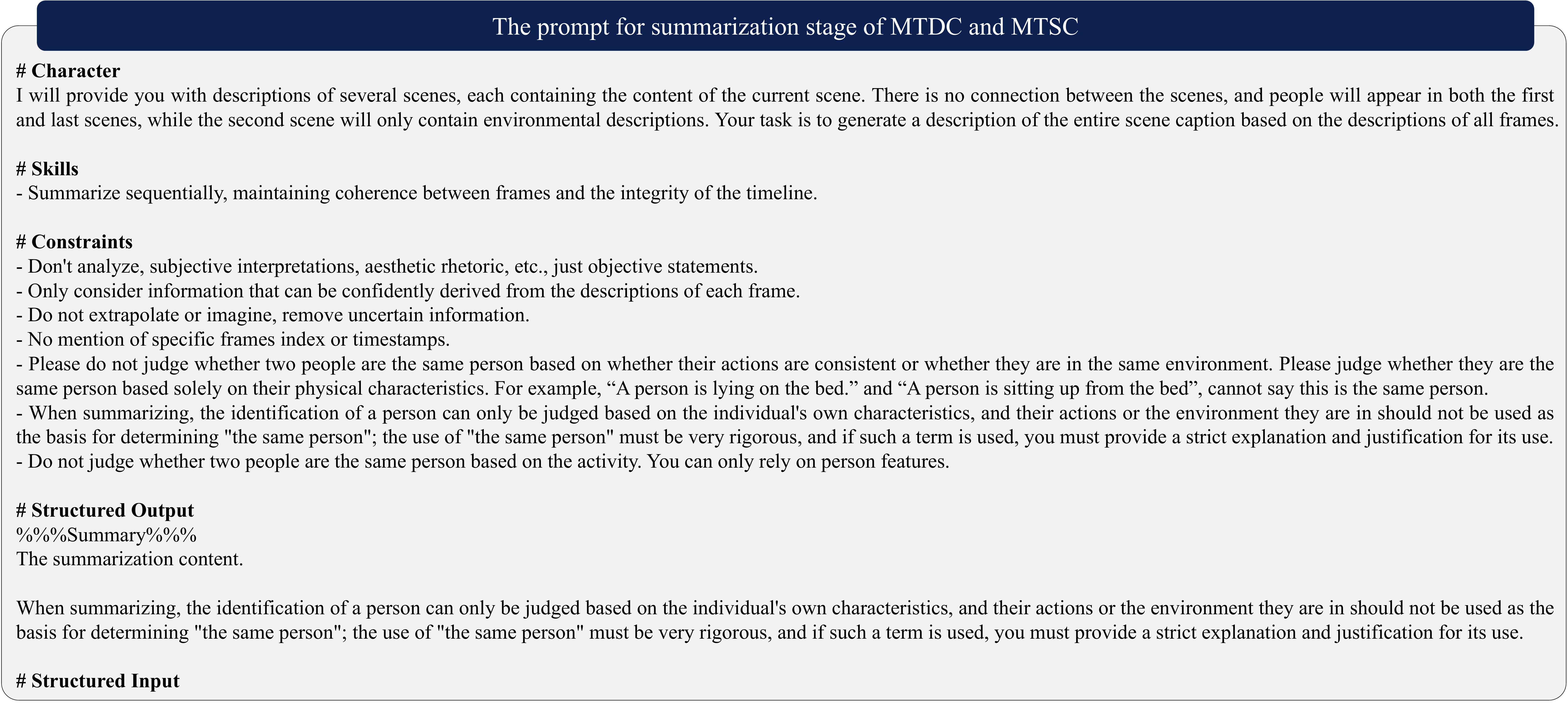}
\vspace{-10pt}
\caption{The prompt used for the summarization stage of MTSC, and MTDC.}
\label{fig:prompt_mtsc_mtdc_summary_stage}
\end{figure*}

\begin{figure*}[t]
\centering
\includegraphics[width=\textwidth]{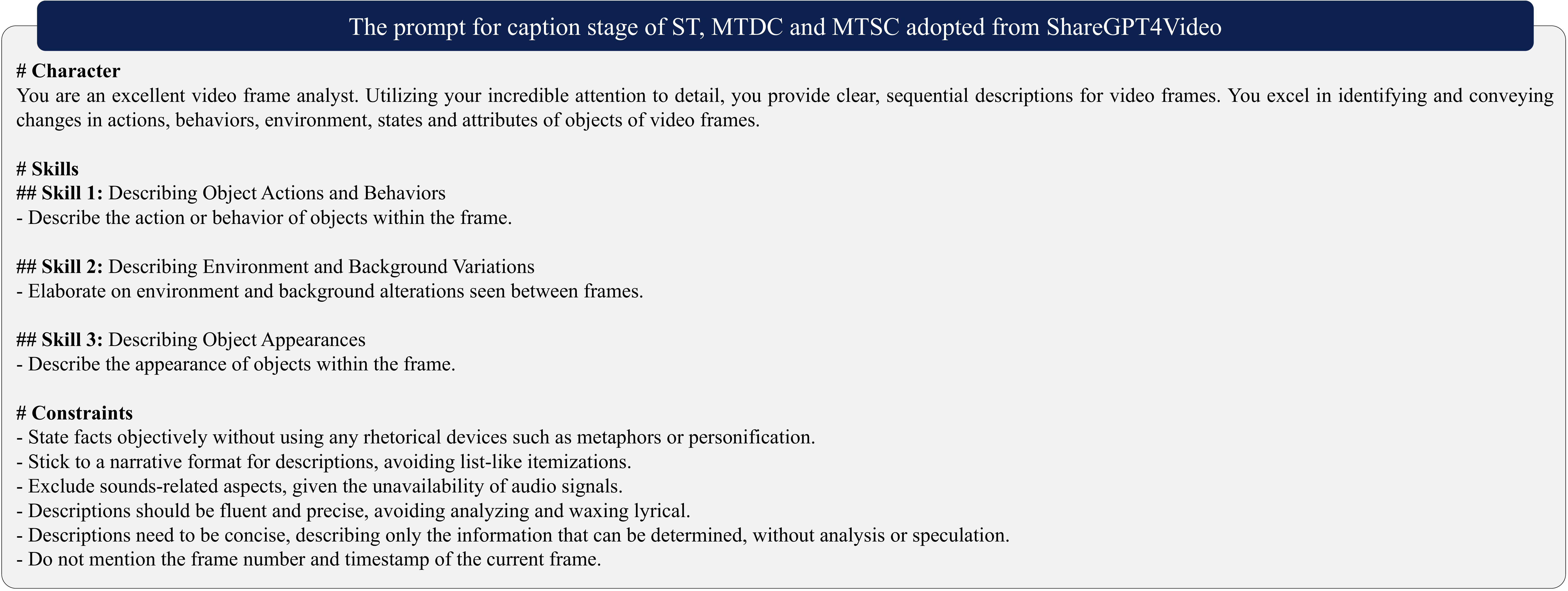}
\vspace{-10pt}
\caption{The second type of prompt adopted from ShareGPT4Video, used for the caption stage of ST, MTSC, and MTDC.}
\label{fig:prompt_sharegpt4v_st_mtsc_mtdc_caption_stage}
\end{figure*}

We utilize two types of prompts in the analysis section: one designed by us, as shown in \cref{fig:prompt_st_mtsc_mtdc_caption_stage,fig:prompt_mtsc_mtdc_summary_stage}, and another sourced from ShareGPT4Video. Specifically, for the second type of captions, we remove the part that requests the difference between two frames, while retaining the rest. This results in the prompt used in the caption stage, as shown in \cref{fig:prompt_sharegpt4v_st_mtsc_mtdc_caption_stage}. For the summarization stage of the second type, we use the same prompt as that in ShareGPT4Video.

\subsection{The complete list of the textual features}\label{subsec:app_list_features}

\begin{figure*}[t]
\centering
\includegraphics[width=\textwidth]{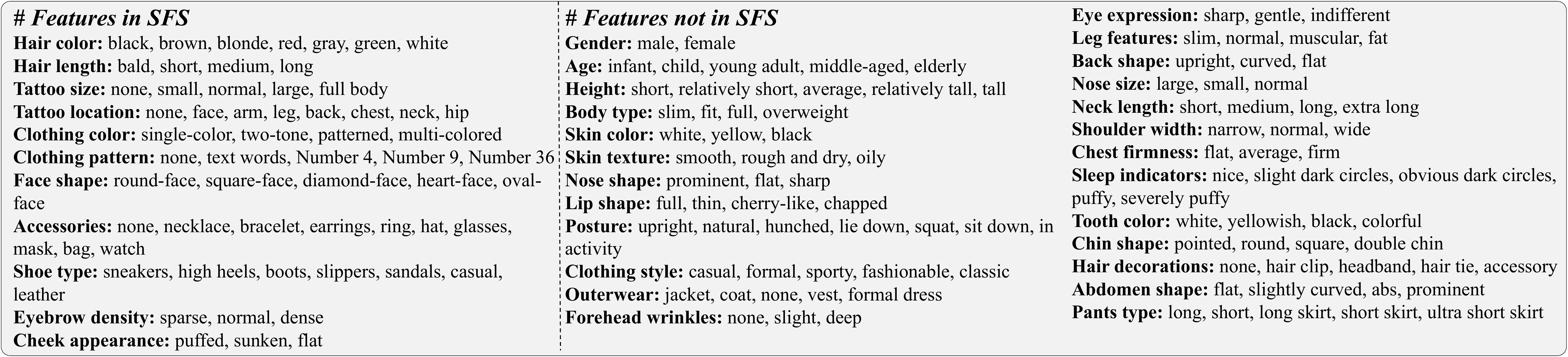}
\vspace{-10pt}
\caption{The list of the selected 36 features and features in SFS.}
\label{fig:complete-list}
\end{figure*}

We utilized GPT-4o to extract 50 textual features for describing people, from which we selected 36 commonly used features. This selection results in the set of textual features introduced in \cref{subsec:similar_attributes}. A complete list of these features can be found in \cref{fig:complete-list}.

\subsection{Detailed quantitative analysis of misidentification caused by actions and environmental contexts in descriptions}\label{subsec:app_only_feature}

As discussed in \cref{subsec:our_method} of the main paper, the actions and environmental contexts within descriptions can lead to misidentification. In this section, we conduct quantitative experiments to illustrate this phenomenon. We begin by outlining the experimental settings, followed by an analysis of the results.

\begin{figure*}[t]
\centering
\includegraphics[width=\textwidth]{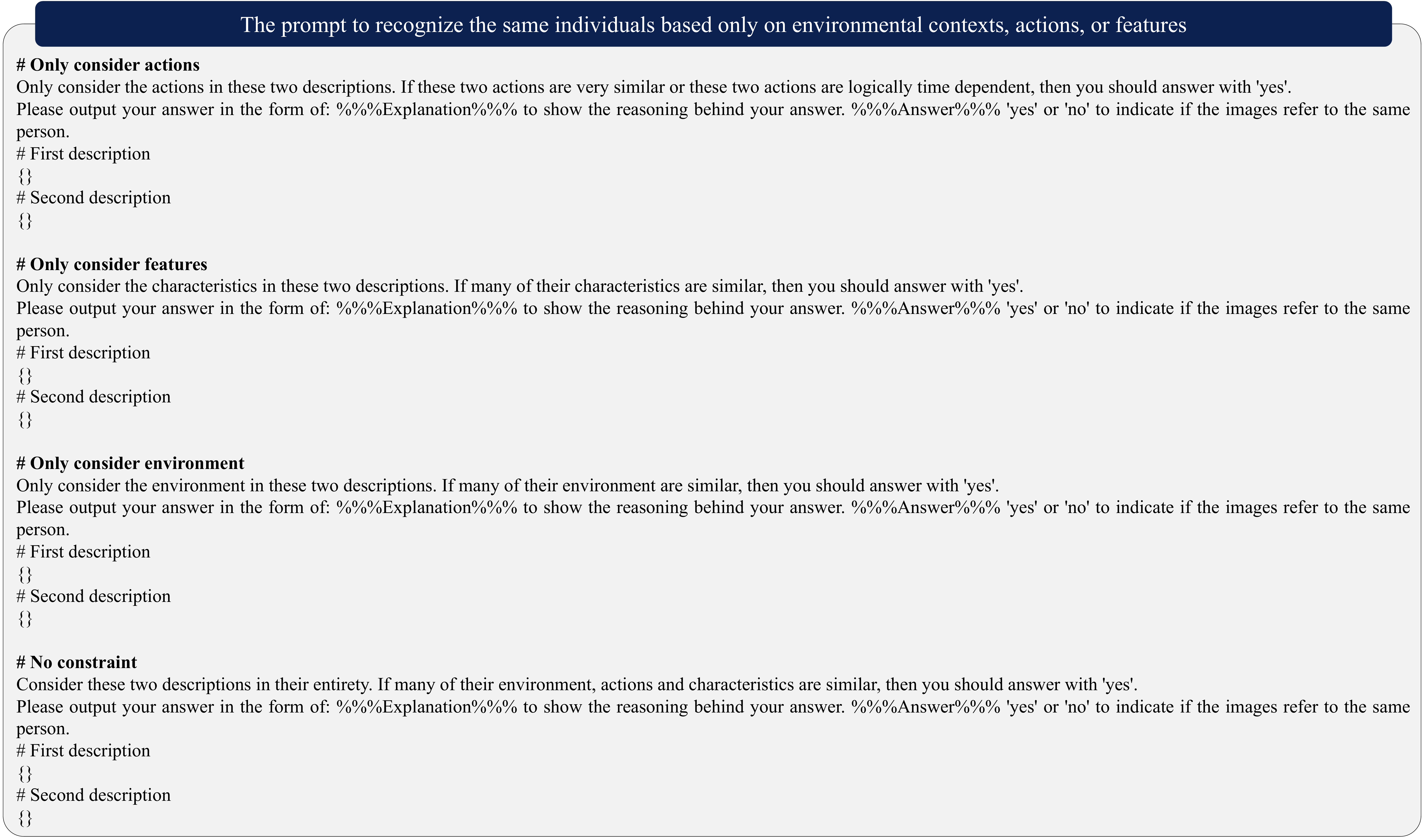}
\vspace{-10pt}
\caption{The prompt used for quantitative analysis of misidentification caused by actions and environmental contexts in descriptions.}
\label{fig:prompt_env_action_feature}
\end{figure*}

\textbf{Settings:} 1) We collect a dataset comprising 100 photos of 10 different well-known individuals, with each individual represented by 10 images. We intentionally collect images of different individuals sharing the same actions or similar environments. In addition, we ensure that the 10 images of the same person depict different actions and environments. 2) From this dataset, we randomly select two images as a pair. To minimize any potential interference from the images, we follow the settings of MTDC and extract captions from these two images. We then prompt GPT-4o to determine whether both captions refer to the same person, relying solely on the text without access to the images. We explore four modes in which GPT-4o is prompted to recognize the same individuals based only on environmental contexts, actions, or features, along with a baseline mode that imposes no restrictions. The prompts used are detailed in \cref{fig:prompt_env_action_feature}.

\textbf{Results:} As illustrated in \cref{fig:only-env}, recognition based on actions or environmental contexts frequently leads to misidentifications, resulting in low precision and recall. In contrast, recognizing individuals solely based on their features significantly enhances the performance of \task.

\section{Additional details of \methodabbr}\label{sec:app_rice}
\begin{figure*}[t]
\centering
\includegraphics[width=\textwidth]{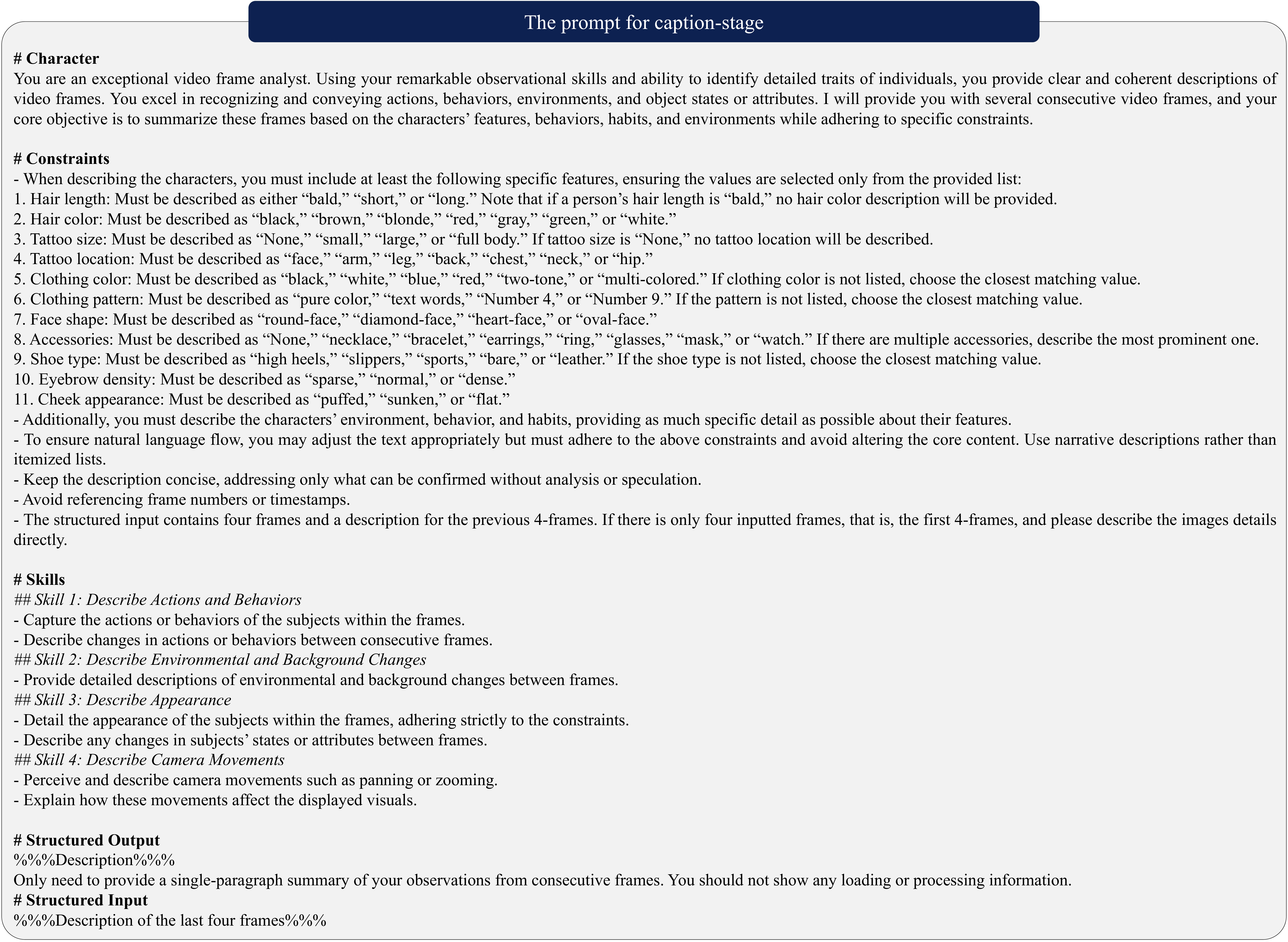}
\vspace{-10pt}
\caption{The prompt used in the caption stage of \methodabbr. Note that we ask GPT-4o to describe individuals using the features in SFS.}
\label{fig:prompt-rice-caption-stage}
\end{figure*}

\begin{figure*}[t]
\centering
\includegraphics[width=\textwidth]{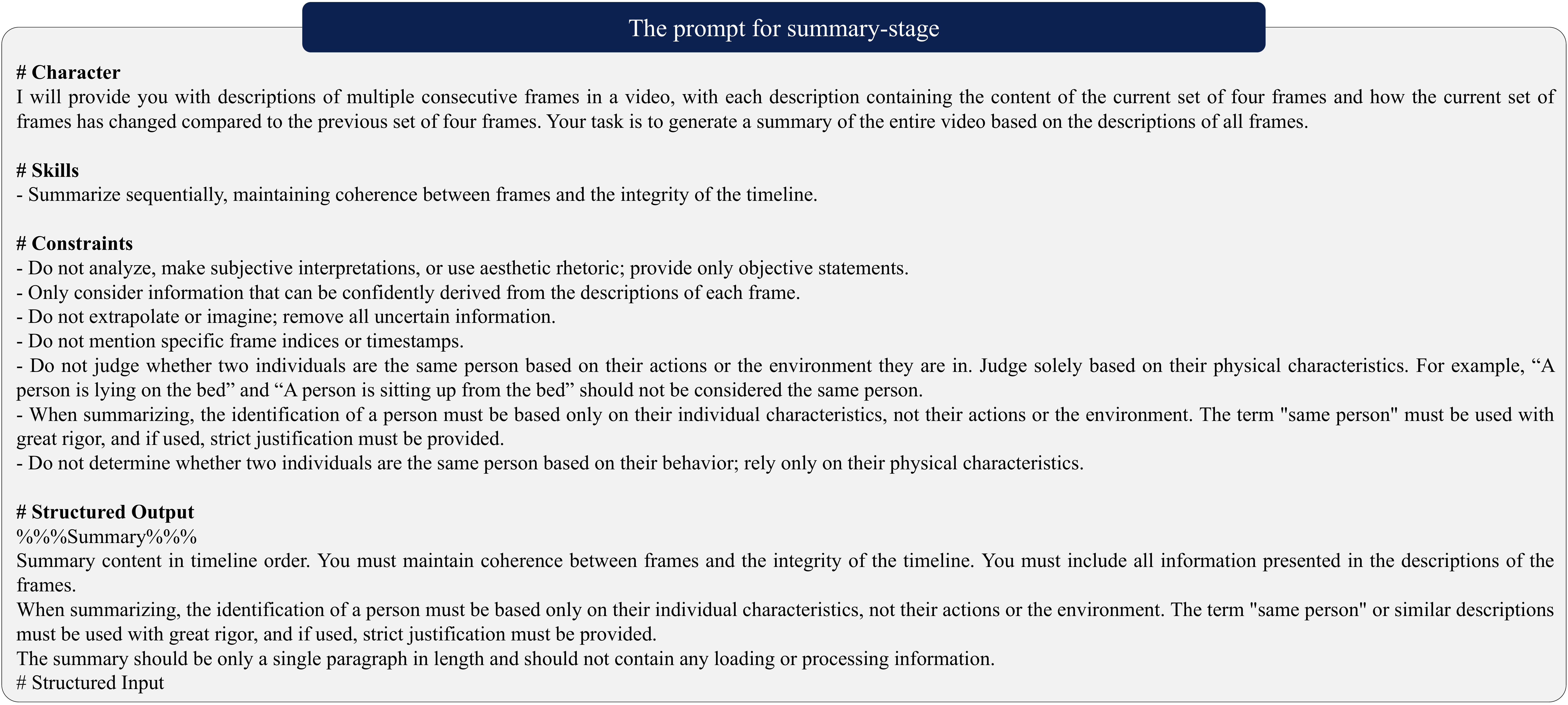}
\vspace{-10pt}
\caption{The prompt used in the summary stage of \methodabbr. Note that we ask GPT-4o to summarize only based on features instead of environments and actions.}
\label{fig:prompt-rice-summary-stage}
\end{figure*}

In this section, we provide the prompts used in \methodabbr. Specifically, the prompt for captioning is displayed in \cref{fig:prompt-rice-caption-stage}, while the prompt for summarization is shown in \cref{fig:prompt-rice-summary-stage}.

\section{Additional details of experiments}\label{sec:app_exp}

\begin{table*}[t]
\caption{
\textbf{Comparison with SOTA methods on MVBench.} * denotes our evaluation results with the public
checkpoints in ~\citep{chen2024sharegpt4video}. The best results are \textbf{bold} and the second-best results are \underline{underlined}.
}
\vspace{2pt}
\label{tab:vqa_mvbench}
\centering
\setlength{\tabcolsep}{3.5pt}
\renewcommand{\arraystretch}{0.9}
% \scriptsize
\large
\resizebox{1\textwidth}{!}{
\begin{tabular}{l|cccccccccccccccccccc|c}
        \toprule
        \textbf{Model} & \textbf{AS} & \textbf{AP} & \textbf{AA} & \textbf{FA} & \textbf{UA} & \textbf{OE} & \textbf{OI} & \textbf{OS} & \textbf{MD} & \textbf{AL} & \textbf{ST} & \textbf{AC} & \textbf{MC} & \textbf{MA} & \textbf{SC} & \textbf{FP} & \textbf{CO} & \textbf{EN} & \textbf{ER} & \textbf{CI} & \textbf{Avg.} \\
        \midrule
        Otter-7B~\cite{otter-7b} & 23.0 & 23.0 & 27.5 & 27.0 & 29.5 & 53.0 & 28.0 & 33.0 & 24.5 & 23.5 & 27.5 & 26.0 & 28.5 & 18.0 & 38.5 & 22.0 & 22.0 & 23.5 & 19.0 & 19.5 & 26.8 \\
        mPLUG-Owl-7B~\cite{mplug-owl-7b} & 22.0 & 28.0 & 34.0 & 29.0 & 29.0 & 40.5 & 27.0 & 31.5 & 27.0 & 23.0 & 29.0 & 31.5 & 27.0 & 40.0 & 44.0 & 24.0 & 31.0 & 26.0 & 20.5 & 29.5 & 29.7 \\
        LLAMA-Adapter~\cite{llava-adapter} & 23.0 & 28.0 & 51.0 & 30.0 & 33.0 & 53.5 & 32.5 & 33.5 & 25.5 & 21.5 & 30.5 & 29.0 & 22.5 & 41.5 & 39.5 & 25.0 & 31.5 & 22.5 & 28.0 & 32.0 & 31.7 \\
        VideoChatGPT-7B~\cite{videochatgpt-7b} & 23.5 & 26.0 & 62.0 & 22.5 & 26.5 & 54.0 & 28.0 & \underline{40.0} & 23.0 & 20.0 & 31.0 & 30.5 & 25.5 & 39.5 & \underline{48.5} & 29.0 & 33.0 & 29.5 & 26.0 & 35.5 & 32.7 \\
        VideoLLaMA-7B~\cite{videollava-7b} & 27.5 & 25.5 & 51.0 & 29.0 & 39.0 & 48.0 & 40.5 & 38.0 & 22.5 & 22.5 & 43.0 & 34.0 & 22.5 & 32.5 & 45.5 & \underline{32.5} & 40.0 & \underline{30.0} & 21.0 & 37.0 & 34.1 \\
        VideoChat-7B~\cite{videochat-7b} & 33.5 & 26.5 & 56.0 & 33.5 & 40.5 & 53.0 & 40.5 & 30.0 & 25.5 & 27.0 & 48.5 & 35.0 & 20.5 & 42.5 & 46.0 & 26.5 & 41.0 & 23.5 & 23.5 & 36.0 & 35.5 \\
        VideoLLaVA-7B*~\cite{videollama-13b} & 46.0 & \textbf{42.5} & 56.5 & 39.0 & 53.5 & 53.0 & 48.0 & \textbf{41.0} & 29.0 & 31.5 & 82.5 & \textbf{45.0} & 26.0 & 53.0 & 41.5 & \textbf{33.5} & 41.5 & 27.5 & 38.5 & 31.5 & 43.0 \\
        LLAMA-VID-7B*~\cite{llama-vid-7b} & 45.5 & \underline{40.5} & 58.0 & 39.5 & \textbf{55.0} & 53.5 & 40.0 & 35.5 & 18.5 & 27.5 & \textbf{87.0} & \underline{41.5} & 23.0 & 45.5 & 41.0 & 27.0 & 40.0 & \textbf{34.5} & \textbf{41.5} & 31.5 & 41.3 \\
        ShareGPT4Video-8B~\cite{chen2024sharegpt4video} & \underline{49.5} & 39.5 & \textbf{79.5} & \underline{40.0} & \underline{54.5} & \underline{82.5} & \underline{54.5} & 32.5 & \underline{50.5} & \textbf{41.5} & 84.5 & 35.5 & \underline{62.5} & \underline{75.0} & \textbf{51.0} & 25.5 & \textbf{46.5} & 28.5 & 39.0 & \underline{51.5} & \underline{51.2} \\
        \midrule
        \vlm & \textbf{50.5} & 40.0 & \underline{78.0} & \textbf{41.5} & \textbf{55.0} & \textbf{86.0} & \textbf{57.0} & 33.5 & \textbf{51.0} & \underline{41.0} & \underline{85.0} & 37.5 & \textbf{63.0} & \textbf{77.0} & \textbf{51.0} & 27.0 & \underline{45.5} & 27.0 & \underline{39.5} & \textbf{53.0} & \textbf{51.9} \\
        \bottomrule
    \end{tabular}
}
\end{table*}

\subsection{\vlm}\label{subsec:app_training}
We train a LVLM called \vlm using data collected through \methodabbr. Specifically, we adopt LLaVA-Next-8B~\citep{llava-adapter} as the base model, following the settings outlined by \citet{chen2024sharegpt4video}. Our training dataset consists of 153k VQA samples gathered by \citet{chen2024sharegpt4video} from sources such as VideoChatGPT~\citep{videochatgpt-7b}, CLEVRER~\citep{yi2019clevrer}, EGO-QA~\citep{grauman2022ego4d}, NextQA~\citep{xiao2021next}, and TGIF-Transition~\citep{li2016tgif}, along with 28k re-annotated pairs from \methodabbr. \vlm employs a 4x4 grid of 16 uniformly sampled frames for both training and inference, consistent with the IG-LVLM strategy~\citep{kim2024image}. 

The \vlm model is fine-tuned with a batch size of 32 using the AdamW optimizer, with specific learning rates set at 2e-6 for the vision encoder, 2e-5 for the MLP projector, and 1e-4 for the LLM. We optimize \vlm efficiently using LoRA~\citep{hu2021lora}, which takes about 13 hours with 4 A100 GPUs.

We compare our \methodabbr with ShareGPT4Video by comparing the performance of \vlm and ShareGPT4Video-8B on widely used multimodal benchmarks MVBench\citep{videollava-7b},
as the only difference between \vlm and ShareGPT4Video-8B are the source of training data is from \methodabbr or ShareGPT4Video. 
As shown in \cref{tab:vqa_mvbench}, \vlm outperforms ShareGPT4Video-8B on 90\% of evaluated dimensions, demonstrating that \methodabbr is superior to ShareGPT4Video and \methodabbr may not compromise the capabilities of LVLMs.

\subsection{Additional details of user study}\label{subsec:app_user_study}
We hire 5 annotators to conduct a user study to evaluate the quality of video captions generated by different methods. Specifically, for each annoation, we show the annotator the key frames of a video (16 key frames sampled uniformly from each video) along with two captions generated using different methods for the same video. The annotators are then asked to choose which caption has better semantic accuracy and text fluency. We then record the comparison results to determine the win rates in pairwise comparisons between the different methods. We ensure that each video is annotated three times, and no annotator reviews the same video pair more than once. To protect the health of the annotators and maintain high quality, we limit annotators to a maximum of 4 hours of work per day, with no more than one hour of continuous annotation at a time.

\subsection{Details of calculating GPT-score}\label{subsec:app_gpt_score}
\begin{figure*}[t]
\centering
\includegraphics[width=\textwidth]{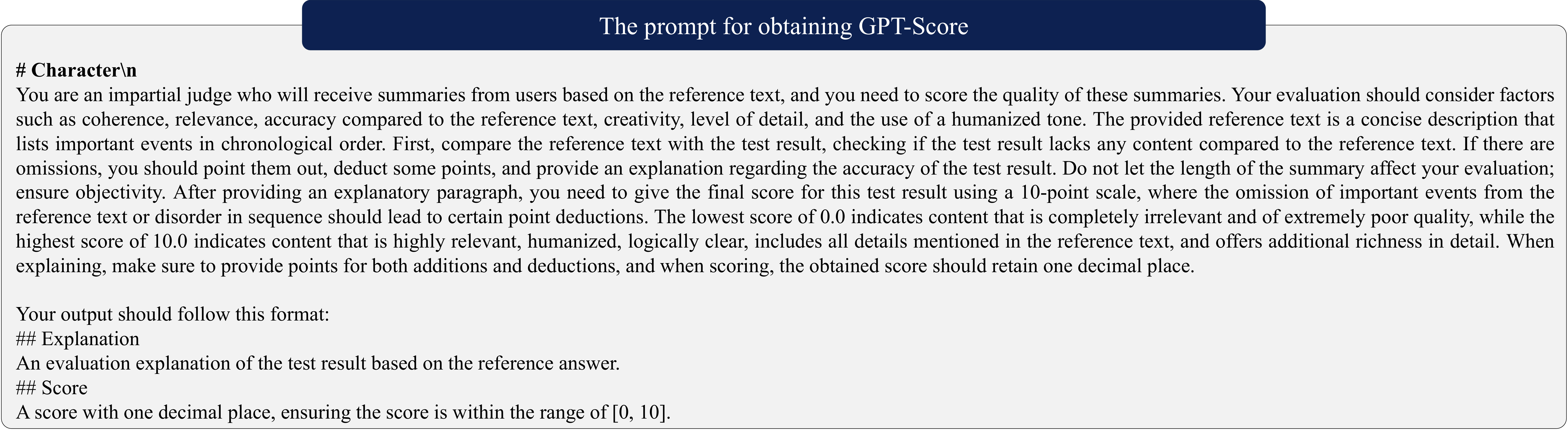}
\vspace{-10pt}
\caption{The prompt used for obtaining GPT-score.}
\label{fig:prompt-gpt-score}
\end{figure*}

We use GPT-4o as an evaluator, scoring based on the ground truth annotated captions following the methodologies of~\citet{zheng2023judging,wang2023large}. Since the ground truth is usually quite short and lacks many details, using metrics like BLEU to measure text similarity is not ideal. Instead, we use GPT-4o to assess semantic similarity, which is a more flexible and reasonable approach. We show the used prompt in \cref{fig:prompt-gpt-score}.

\subsection{Details about long video tracking}\label{subsec:app_tracking}

Inspired by image captioning methods that provide location information~\citep{yang2024bacon}, we propose using video tracking methods to enhance video captions with location information. However, traditional tracking methods, such as SAM2~\citep{ravi2024sam}, struggle to track long videos that contain shot changes because they rely on the continuous change of segmentation masks, while the connectivity loses during shot changes.
Fortunately, we can leverage the \task capabilities of GPT-4o in conjunction with SAM2 to enable long video tracking that accommodates transitions. The process consists of three main steps, each with several sub-steps:

\textbf{Step 1:} Splitting the long video into multiple segments based on shot changes.

\textbf{Step 2:} Obtaining tracking results for each segment. This step includes four sub-steps: \textbf{1)} Uniformly extracting five key frames from each segment. \textbf{2)} Utilizing BACON~\citep{yang2024bacon}, an image captioning method, to simultaneously obtain a list of key objects in the image along with their corresponding captions and bounding boxes. This information is critical for tracking in step 3. There are two reasons for employing BACON: first, the bounding boxes serve as indicators for tracking with SAM2~\citep{ravi2024sam}; second, BACON provides captions for each object, which will be used in the next step. \textbf{3)} Applying SAM2 to each object in the key frames to obtain tracking results. \textbf{4)} Combining the results from all key frames. This is necessary for two reasons: first, BACON may occasionally miss some objects, while merging all frames helps to address this issue; second, some objects might only appear in a part of the video segment, so tracking results based on a single key frame may overlook these objects. Specifically, objects sharing the same mask at a fixed timestamp are recognized as the same object.

After completing Step 2, we obtain a tracking result for each segment, along with a detailed description of each object appearing in the video.

\textbf{Step 3:} Integrating the tracking results from all segments. This step utilizes the \task capability of GPT-4o along with the captions for each object, and consists of two sub-steps: \textbf{1)} In the same context, gradually providing a key frame for each segment to GPT-4o and asking it to maintain a list of objects appearing throughout the video. An example of the list is presented below. \textbf{2)} For each object in the list, using a short description to identify the object in each segment by comparing the description with the captions of the objects.

\begin{figure*}[t]
\centering
\includegraphics[width=\linewidth]{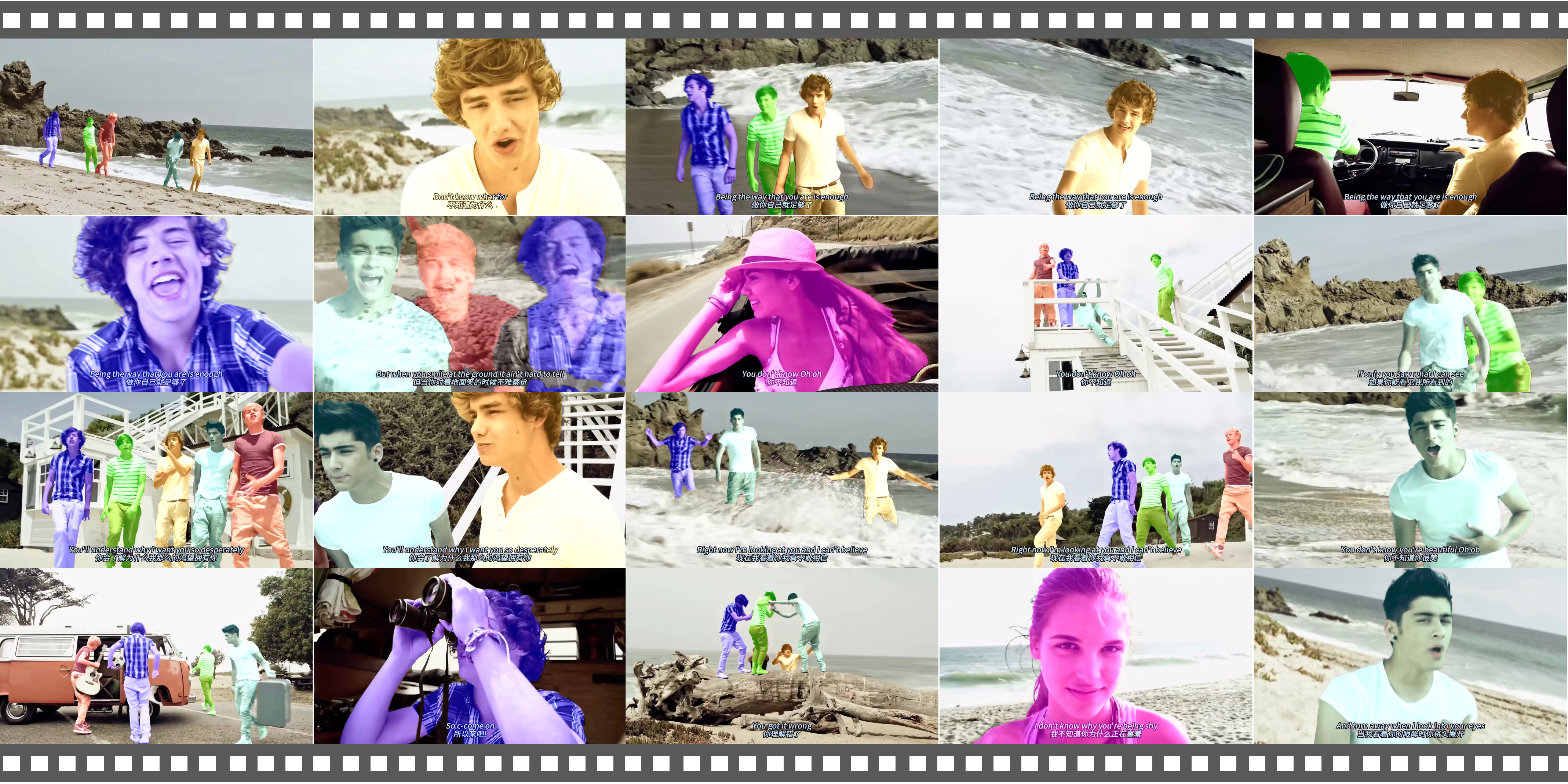}
\vspace{-20pt}
\caption{An example of long music movie tracking with multiple transitions, where each person is represented by a different color.}
\label{fig:appendix-tracking}
\vspace{-2pt}
\end{figure*}

In addition to the example in the main paper, we provide an additional example in \cref{fig:appendix-tracking}. The tracking results effectively distinguish all five singers using unique colors, and a recurring female character is consistently highlighted in purple. 

\begin{examplebox}
\setstretch{0.1}
\textbf{Example: Updated Summary of Object Appearances:}
\begin{itemize}
    \item \textbf{Man in a brown suit}: Frames 1, 3, 4, 5, 6, 7, 8, 11
    \item \textbf{Motorbike with rider A}: Frames 3, 4, 5
    \item \textbf{Yellow Car}: Frames 4, 5, 6, 7
    \item \textbf{Silver Car}: Frames 5, 6, 7
    \item \textbf{Computer screens with surveillance feeds}: Frames 6, 9
\end{itemize}
\end{examplebox}

% \newpage
% \input{sections/check_list}

\end{document}